\documentclass[10pt]{article} 
\usepackage[accepted]{tmlr}

\usepackage{amsmath,amsfonts,bm}

\def\eqref#1{equation~\ref{#1}}

\def\1{\bm{1}}

\def\ra{{\textnormal{a}}}

\def\rx{{\textnormal{x}}}

\def\rva{{\mathbf{a}}}

\def\erva{{\textnormal{a}}}

\def\ervx{{\textnormal{x}}}

\def\rmA{{\mathbf{A}}}

\def\vmu{{\bm{\mu}}}
\def\vtheta{{\bm{\theta}}}
\def\va{{\bm{a}}}

\def\ve{{\bm{e}}}

\def\vx{{\bm{x}}}

\def\eva{{a}}

\def\mA{{\bm{A}}}

\def\mH{{\bm{H}}}
\def\mI{{\bm{I}}}
\def\mJ{{\bm{J}}}

\def\mX{{\bm{X}}}

\def\mSigma{{\bm{\Sigma}}}

\DeclareMathAlphabet{\mathsfit}{\encodingdefault}{\sfdefault}{m}{sl}
\SetMathAlphabet{\mathsfit}{bold}{\encodingdefault}{\sfdefault}{bx}{n}
\newcommand{\tens}[1]{\bm{\mathsfit{#1}}}
\def\tA{{\tens{A}}}

\def\tX{{\tens{X}}}

\def\gG{{\mathcal{G}}}

\def\sA{{\mathbb{A}}}
\def\sB{{\mathbb{B}}}

\def\sS{{\mathbb{S}}}

\def\emA{{A}}

\newcommand{\etens}[1]{\mathsfit{#1}}

\def\etA{{\etens{A}}}

\newcommand{\E}{\mathbb{E}}

\newcommand{\R}{\mathbb{R}}

\newcommand{\KL}{D_{\mathrm{KL}}}
\newcommand{\Var}{\mathrm{Var}}

\newcommand{\Cov}{\mathrm{Cov}}

\newcommand{\normltwo}{L^2}
\newcommand{\normlp}{L^p}

\newcommand{\parents}{Pa} 

\DeclareMathOperator*{\argmin}{arg\,min}

\usepackage{hyperref}
\usepackage{url}
\usepackage{graphicx} 
\usepackage{amsmath}
\usepackage{graphicx}
\usepackage{subcaption}
\usepackage{amsmath}
\usepackage{wrapfig} 
\usepackage{amssymb}
\usepackage{booktabs}
\usepackage{amssymb}
\usepackage{pifont}
\usepackage{multirow}
\usepackage{wrapfig}
\usepackage{algorithm,algpseudocode}
\usepackage{amsmath}
\usepackage{array}
\usepackage{wrapfig}  
\usepackage{amsthm}
\usepackage{booktabs}
\usepackage{enumitem}
\usepackage{graphicx}
\usepackage{color}
\usepackage{algorithm,algpseudocode}
\usepackage{natbib}  
\usepackage{caption} 
\usepackage{algorithm,algpseudocode}
\usepackage{newfloat}
\usepackage{listings}

\title{FedDUAL: A Dual-Strategy with Adaptive Loss and Dynamic Aggregation for Mitigating Data Heterogeneity in Federated Learning}

\author{
  \name Pranab Sahoo \email pranab\_2021cs25@iitp.ac.in \\
  \addr Department of Computer Science\\
  Indian Institute of Technology Patna, India
  \AND
  \name Ashutosh Tripathi \email ashutoshtripathi191@gmail.com \\
  \addr Electrical and Electronics Engg. Department\\ Rajiv Gandhi Institute of Petroleum Technology, India
  \AND
  \name Sriparna Saha \email sriparna@iitp.ac.in \\
  \addr Department of Computer Science\\
  Indian Institute of Technology Patna, India
  \AND
  \name Samrat Mondal \email samrat@iitp.ac.in \\
  \addr Department of Computer Science\\
  Indian Institute of Technology Patna, India}

\def\month{11}  
\def\year{2025} 
\def\openreview{\url{https://openreview.net/forum?id=8M3XfmNhTZ}} 

\begin{document}

\maketitle

\maketitle

\begin{abstract}Federated Learning (FL) marks a transformative approach to distributed model training by combining locally optimized models from various clients into a unified global model. While FL preserves data privacy by eliminating centralized storage, it encounters significant challenges such as performance degradation, slower convergence, and reduced robustness of the global model due to the heterogeneity in client data distributions. Among the various forms of data heterogeneity, label skew emerges as a particularly formidable and prevalent issue, especially in domains such as image classification. To address these challenges, we begin with comprehensive experiments to pinpoint the underlying issues in the FL training process, such as gradient instability and the emergence of sharp minima in the global model, both of which contribute to performance inconsistencies. Based on our findings, we introduce an innovative dual-strategy approach designed to effectively resolve these issues. First, we introduce an adaptive loss function for client-side training, meticulously crafted to preserve previously acquired knowledge while maintaining an optimal equilibrium between local optimization and global model coherence. Secondly, we develop a dynamic aggregation strategy for aggregating client models at the server. This approach adapts to each client's unique learning patterns, effectively addressing the challenges of diverse data across the network. Our comprehensive evaluation, conducted across three diverse real-world datasets, coupled with theoretical convergence guarantees, demonstrates the superior efficacy of our method compared to several established state-of-the-art approaches. The code can be found at~\url{https://github.com/Pranabiitp/FedDUAL}.

\end{abstract}


\section{Introduction}
Federated learning (FL) has revolutionized collaborative model training by enabling multiple clients to contribute to a global model without compromising the privacy of their local data~\citep{mcmahan2017communication}. This decentralized strategy avoids the need for sending data to a central server, thus maintaining data privacy. As the digital landscape evolves, with an increasing number of distributed data sources emerging from mobile devices, healthcare institutions, and Internet of Things (IoT) networks, FL has emerged as a pivotal solution for training sophisticated deep networks across geographically dispersed and heterogeneous environments~\citep{bonawitz2016practical},~\cite{sahoo2024adafedprox},~\citep{hu2024fedmut}. However, a significant practical obstacle encountered during federated training is data heterogeneity in the form of skewness in labels and quantity of the data across various clients~\citep{kairouz2021advances},~\citep{li2020federated}. Diverse user behaviors can lead to significant heterogeneity in the local data of different clients, leading to non-independent and identically distributed (non-IID) data. This variability can introduce biases in model training, leading to unstable convergence and potentially degrading the model’s performance or making it counterproductive~\citep{li2022federated},~\citep{zhao2018federated}. While FedAvg~\citep{mcmahan2017communication} is effective and widely used, it often falls short in accuracy and convergence with static aggregation methods. These methods combine model updates from different clients in a fixed manner, failing to adapt to heterogeneous data distributions and client drift, as discussed in~\citep{karimireddy2020scaffold}.

Previous studies have addressed the issue of client drift by implementing penalties for deviations between client and server models~\citep{li2020federated},~\citep{li2021model}, employing variance reduction techniques during client updates~\citep{karimireddy2020scaffold},~\citep{acar2021federated}, or utilizing novel aggregation methods on the server side~\citep{Chen_2023_CVPR},~\citep{chowdhury2024fedsatstatisticalaggregationapproach}.

\begin{wrapfigure}{r}{0.4\textwidth} 
    \centering
    \includegraphics[width=0.44\textwidth]{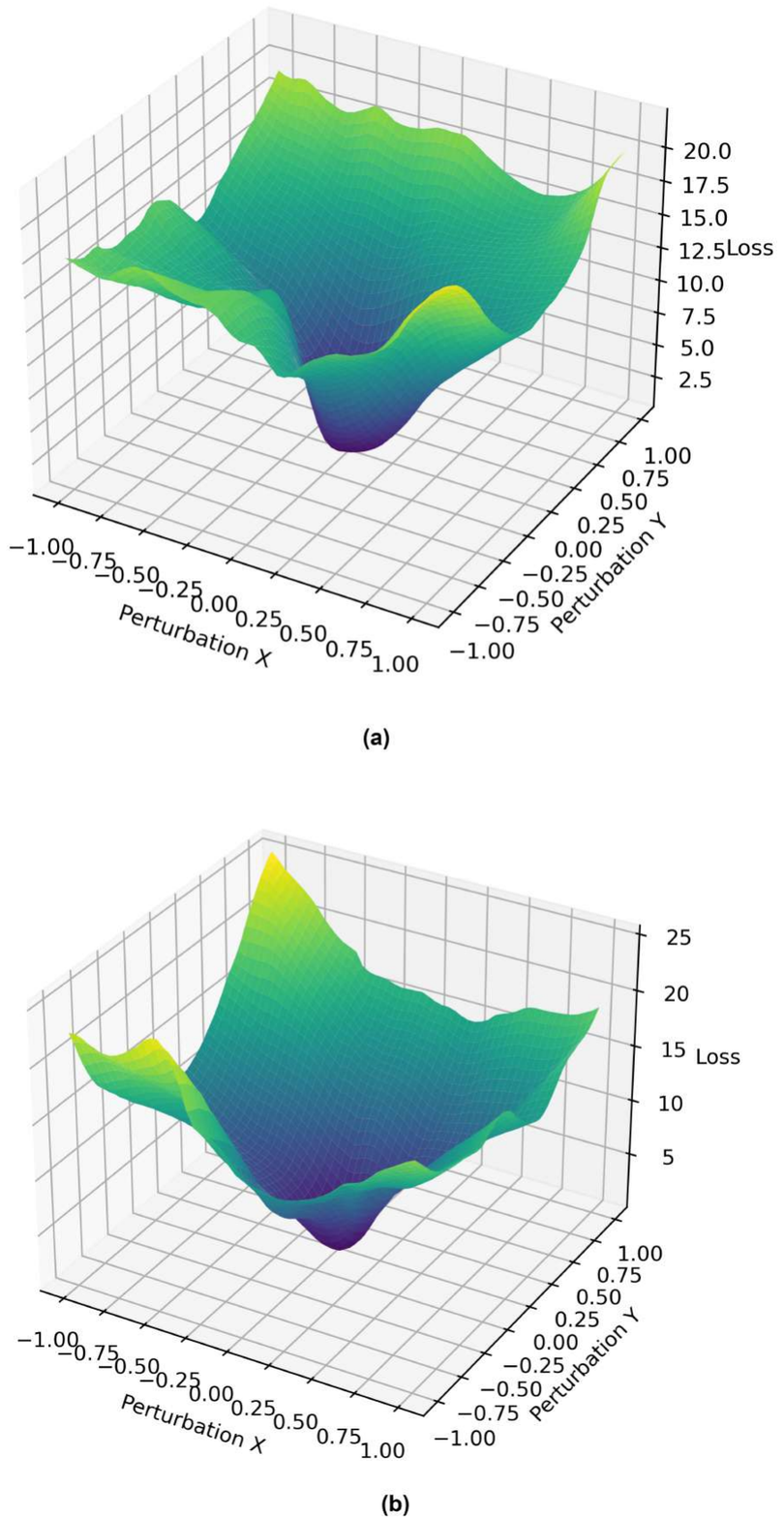} 
    \caption{Visualization of the loss surface for the global model trained on the FMNIST dataset using the FedAvg algorithm: (a) depicts the loss landscape when trained on IID data, while (b) illustrates the landscape for non-IID data distribution.}
    \label{unixi}
\end{wrapfigure}

\subsection{Motivation}
\label{motivation}
\begin{figure*}[htbp]
    \centering
    \begin{subfigure}[t]{0.48\textwidth}
        \centering
        \includegraphics[width=\linewidth]{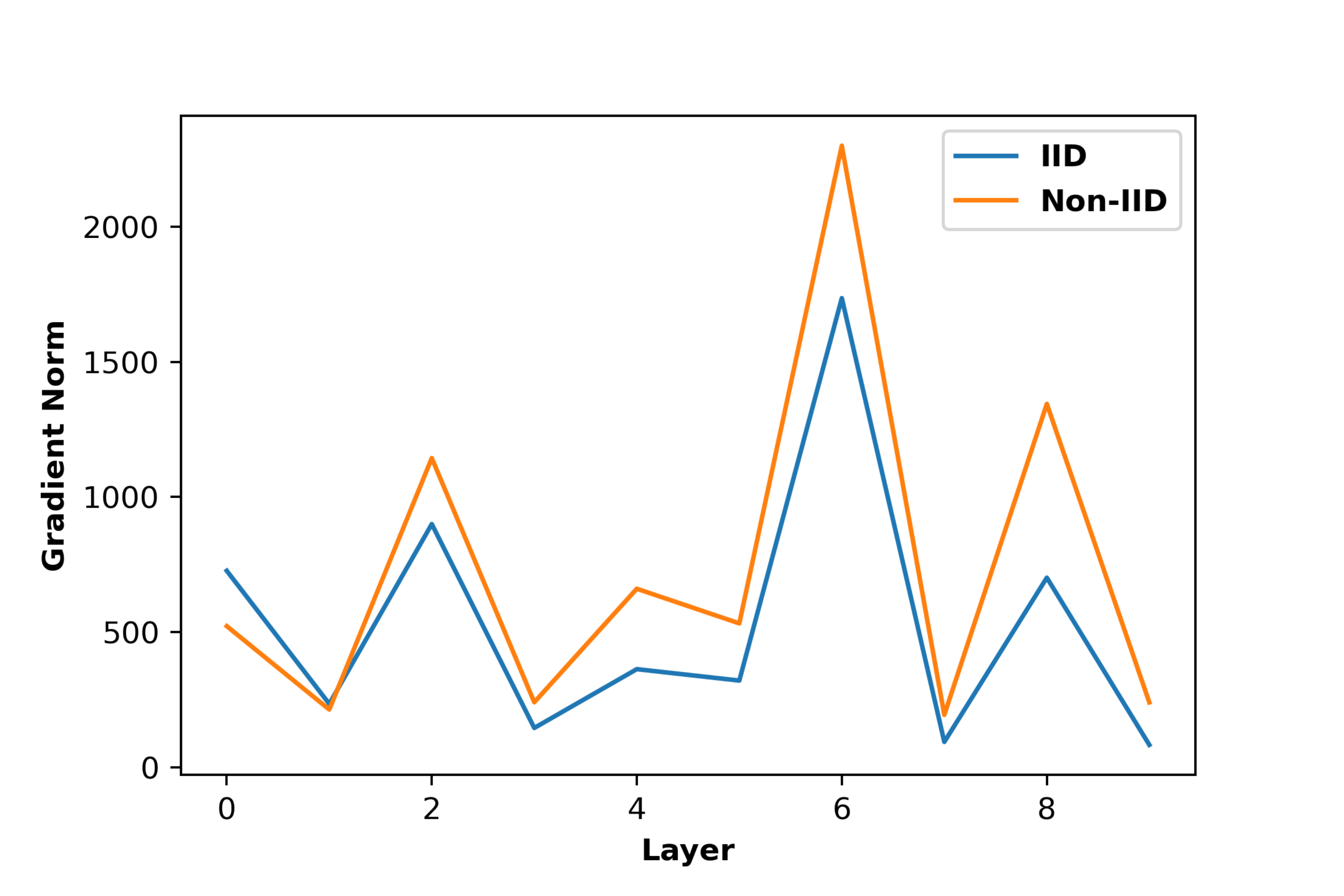}
        \caption{}
        \label{fig21}
    \end{subfigure}
    \hfill
    \begin{subfigure}[t]{0.50\textwidth}
        \centering
         \includegraphics[height=0.21\textheight]{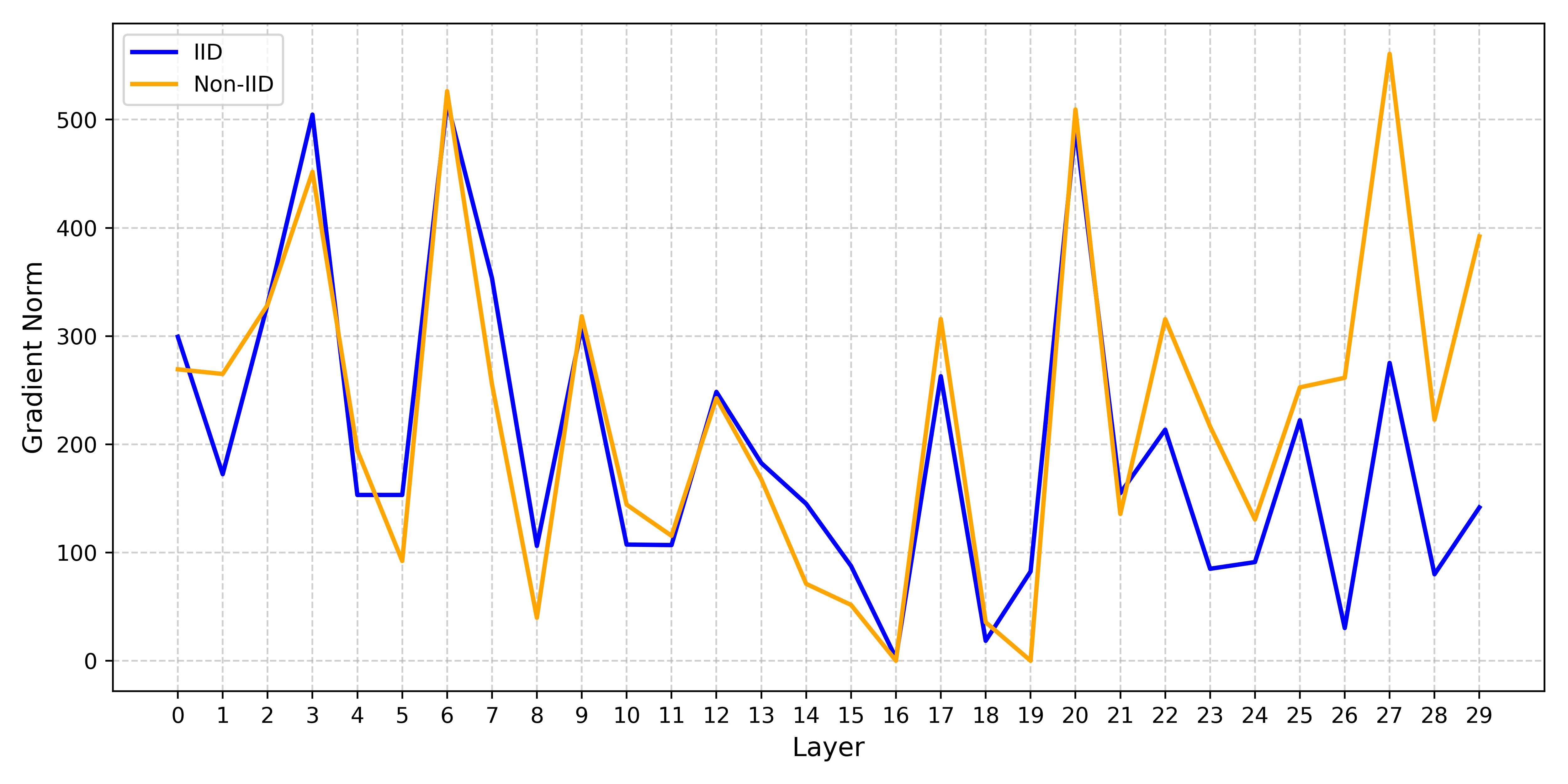}
        \caption{}
        \label{zxcvb}
    \end{subfigure}

    \caption{Comparison of gradient norms between models trained on IID and non-IID datasets using the FedAvg algorithm. (a) FMNIST dataset using LeNet model. (b) CIFAR10 dataset using VGG16 model.}
    \label{fig:grad_norm_comparison}
\end{figure*}

Prior studies by~\cite{Yashwanth_2024_WACV},~\cite{hu2024fedmut} have demonstrated that in non-IID scenarios, federated models tend to converge to 'sharp minima', resulting in significant performance degradation and compromised generalizability. In this study, we investigate the root causes of this phenomenon and propose a novel solution to mitigate its effects. Our study begins with a detailed analysis of loss landscapes for FedAvg-trained models across IID and non-IID data distributions. Figure~\ref{unixi} visually depicts the loss landscapes of two models on the FMNIST dataset with systematic parameter perturbations. The model trained on IID data exhibits a notably smoother and wider valley in its loss landscape, suggesting greater robustness and better generalization. In contrast, the model trained on non-IID manifests sharper peaks and narrower valleys, indicating higher sensitivity to parameter variations and potential overfitting. These visualizations offer strong evidence that in the presence of non-IID data, the FedAvg algorithm achieves suboptimal generalization. Motivated by this observation, we investigate the underlying mechanisms by analyzing gradient norms to identify which parts of the neural network are most affected by data heterogeneity. Our findings, presented in Fig.~\ref{fig:grad_norm_comparison} reveal a notable pattern: in non-IID scenarios, the gradient norms of the final layers, including the classification layer, exhibit significant amplification compared to their IID counterparts. Such amplification leads to model instability, impedes convergence, and ultimately compromises the generalizability of the federated model. Our investigation suggests that effective federated training in non-IID environments necessitates targeted adjustments during server-side aggregation, particularly for these highly affected layers, to achieve performance comparable to IID settings. 

This prompts one critical question: Can static aggregation methods effectively address severe non-IID data distributions across clients while maintaining higher convergence, performance, and generalizability in federated models? The answer is decidedly negative. Static aggregation methods inherently struggle with the dynamic heterogeneity present in federated networks, where adjusting parameters based on client distributions and performance in each communication round is crucial. Although incorporating predetermined parameters into the aggregation process may provide some partial mitigation, these methods fail to address the complex challenges posed by non-IID data distributions. A more dynamic and nuanced approach is necessary to effectively manage these multifaceted issues. To address this challenge, we apply dynamic aggregation to the model's final layers, where gradient norms fluctuate significantly in non-IID scenarios, while using traditional aggregation (FedAvg) for the lower layers. For dynamic aggregation, we leverage the concept of Wasserstein Barycenter~\citep{agueh2011barycenters}, derived from optimal transport theory, to integrate client-specific learning behaviors in these affected layers. By minimizing discrepancies from non-IID data, the Wasserstein Barycenter helps to align gradients from diverse clients, offering precise model updates. This approach ensures fair aggregation, adapts to data heterogeneity, reduces bias, and enhances robustness, ultimately leading to more stable model convergence and improved generalization.



In addition to the server-side dynamic aggregation, we introduce an adaptive loss function for local training on the client side. This function allows clients to effectively explore the minima on their local datasets while preventing overfitting, thereby enhancing local optimization. Simultaneously, it preserves the global knowledge of the federated model, ensuring that the benefits from all participating clients are integrated. By incorporating a regularization parameter, $\beta$, the local loss function dynamically balances the trade-offs between local and global objectives. The contributions of this paper are as follows:


\begin{itemize}
\item We introduce FedDUAL, an innovative dual-strategy approach designed to effectively develop a robust and generalized federated model in highly heterogeneous data environments.


\item We introduce an adaptive loss function for client-side training to balance the trade-offs between local and global objectives.
\item Instead of straightforward server-side averaging, we propose a dynamic aggregation technique that uses Wasserstein Barycenter to reduce the effects of non-IID data by integrating the learning behaviors of participating clients.

\item We conducted extensive experiments on three real-world datasets, demonstrating significant performance improvements over state-of-the-art methods and offering theoretical convergence guarantees for both convex and non-convex scenarios.

\end{itemize}


\section{Related Work}
The landscape of FL research has been significantly shaped by efforts to address data heterogeneity challenges, yielding a diverse array of innovative solutions. These approaches can be divided into three primary categories: (1) client drift mitigation strategies, which refine local client objectives to foster better alignment with the global model ~\citep{li2021model},~\citep{karimireddy2020scaffold},~\citep{acar2021federated},~\citep{luo2021no},~\citep{li2023effectiveness} (2) aggregation scheme optimization, aimed at enhancing server-side fusion of model updates~\citep{hsu2019measuring},~\citep{lin2020ensemble},~\citep{wang2020tackling},~\citep{wang2020federated} and (3) personalized FL, which tailors models to individual clients~\citep{fallah2020personalized},~\citep{sattler2020clustered},~\citep{bui2019federated}. Our research primarily focuses on two interconnected aspects of FL: mitigating client drift and optimizing server-side aggregation, and we will discuss the same in the literature review.

\cite{mcmahan2017communication} introduced FL as an extension of local Stochastic Gradient Descent (SGD)~\citep{stich2019localsgdconvergesfast}, enabling increased local gradient updates on client devices before server synchronization and significantly reducing communication costs in identically distributed data settings. However, the method faces considerable obstacles when dealing with non-IID scenarios. Since then, various methods have emerged to address the challenge of data heterogeneity in FL~\citep{li2019convergence},~\citep{yang2021achieving},~\citep{lin2018don},~\citep{hsu2019measuring}. FedProx~\citep{li2020federated} incorporates a proximal regularization term to the optimization function to reduce model drift and addresses client stragglers. However, this term can also lead to local updates being biased towards the previous global model, which may result in misalignment between local and global optima.
Building on previous work,~\citet{acar2021federated} introduced a dynamic regularization term to align local updates more closely with global model parameters, effectively reducing client drift caused by local model overfitting.~\cite{sun2023efficient} further advanced the field with a momentum-based algorithm that accelerates convergence by combining global gradient descent with a locally adaptive optimizer. Similarly, several studies use variance reduction techniques, such as SCAFFOLD~\citep{karimireddy2020scaffold}. However, this approach often results in higher communication costs due to the transmission of additional control variates~\citep{halgamuge2009estimation}. FedPVR~\citep{li2023effectiveness} addresses these limitations by reassessing FedAvg's performance on deep neural networks, uncovering substantial diversity in the final classification layers. By proposing a targeted variance reduction strategy focused solely on these final layers, FedPVR outperforms several benchmarks. MOON~\citep{li2021model} introduces an innovative model-contrastive framework leveraging a contrastive loss to align local client representations with the global model, effectively mitigating client drift, and enhancing convergence, particularly in challenging non-IID environments.~\cite{luo2021no} introduced CCVR (Classifier Calibration and Variance Reduction), which employs a classifier regularization and calibration method to enhance federated learning performance. CCVR's approach involves fine-tuning the classifier using virtual representations sampled from an approximated Gaussian mixture model.~\cite{shi2023make} introduced a novel differentially private federated learning (DPFL) algorithm that integrates the Sharpness-Aware Minimization (SAM) optimizer to enhance stability and robustness against weight perturbations. By generating flatter loss landscapes and reducing the impact of differential privacy (DP) noise, it mitigates performance degradation and achieves state-of-the-art results, supported by theoretical analysis and rigorous privacy guarantees.~\cite{fani2024accelerating} proposed FED3R, leveraging Ridge Regression on pre-trained features to tackle non-IID data challenges, effectively mitigating client drift, enhancing convergence, and optimizing efficiency in cross-device settings.

Another line of research targets optimizing server-side aggregation in FL. For instance,~\cite{hsu2019measuring} investigated the impact of non-IID data on visual classification by creating datasets with diverse distributions and found that increased data heterogeneity negatively affected performance, leading them to propose server momentum as a potential solution. FedNova~\citep{wang2020tackling} addressed the problem of objective inconsistency due to client heterogeneity in federated optimization by introducing a normalized averaging technique, which resolves this inconsistency and ensures rapid error convergence. Addressing the limitations of traditional parameter averaging methods,~\cite{lin2020ensemble} introduced ensemble distillation for model fusion. This approach allows for the flexible aggregation of heterogeneous client models by training a central classifier on unlabeled data, using the outputs from the client models as guidance. FedMRL~\citep{sahoo2024fedmrldataheterogeneityaware} introduced a novel framework by using a loss function that promotes fairness among clients and employed a multi-agent reinforcement learning for personalized proximal terms , and a self-organizing map to dynamically adjust server-side weights during aggregation.

\section{Definitions}
\label{subsec:notations}

In this subsection, we summarize the key mathematical definitions used in the paper to ensure clarity.

\paragraph{Kullback–Leibler (KL) Divergence.} 
The KL divergence measures how one probability distribution diverges from a second reference distribution. For two distributions $P$ and $Q$ over the same probability space, it is defined as:
\begin{equation}
D_{\mathrm{KL}}(P \,\|\, Q) = \sum_{i} P(i) \log \frac{P(i)}{Q(i)}.
\end{equation}

\paragraph{Wasserstein Distance.} 
The Wasserstein distance (also known as the Earth Mover’s Distance) measures the optimal cost of transporting mass to transform one probability distribution into another. For two distributions $\mu$ and $\nu$, the $p$-Wasserstein distance is defined as:
\begin{equation}
W_p(\mu, \nu) = \left( \inf_{\gamma \in \Gamma(\mu,\nu)} 
\int_{\mathcal{X}\times\mathcal{X}} d(x,y)^p \, d\gamma(x,y) \right)^{1/p},
\end{equation}
where $\Gamma(\mu,\nu)$ denotes the set of all couplings with marginals $\mu$ and $\nu$, and $d(\cdot,\cdot)$ is a ground metric. 
We leverage the Wasserstein distance and its barycenter to aggregate client models in a manner that accounts for data heterogeneity.

\paragraph{Wasserstein Barycenter.}
Given distributions $\{\mu_k\}_{k=1}^K$ and weights $\{\lambda_k\}_{k=1}^K$, the Wasserstein barycenter $\hat{\mu}$ is defined as the distribution minimizing the weighted sum of Wasserstein distances:
\begin{equation}
\hat{\mu} = \arg\min_{\nu} \sum_{k=1}^K \lambda_k W_p(\mu_k, \nu).
\end{equation}
This barycenter allows us to aggregate the last-layer representations of client models more effectively than simple averaging, improving robustness to non-IID data.

\section{Methods and Materials}
We consider a practical FL scenario with non-IID data distribution among $K$ independent clients, each with local training data $D_k(x, y)$, where $(x,y)$ denoting the data points. We initialize the global model weights $\theta_r^g$ and share it to the participating clients. 
The clients download the weighs from the server and train it using their local dataset $D_k(x, y)$. The updated model parameters $\theta^r_k$ from each client $k$ for $r^{th}$ communication round are uploaded to the server to aggregate into a global model $\theta_r^g$. Our objective is to develop a robust global model by collaboratively training local models across clients, even under varying heterogeneous conditions. To formalize, we define the optimal global model \( \theta^* \) as follows:


\begin{equation}
\begin{aligned}
    \theta^* := \argmin_{\theta} F(\theta),  \hspace{2pt}F(\theta) := \frac{1}{K} \sum_{k} f_k(\theta),
\end{aligned}
\end{equation}

where $f_k(\theta)$ is defined in Eq.~\ref{eq11c}.

\begin{equation}
\label{eq11c}
{f}_k(\theta) = {E}_{(x,y) \sim D_k} [\ell(f_{\theta}(x), y)],
\end{equation}

where $\theta$ represents the global model parameters,  $f_{\theta}(x)$ is the model's prediction, and $\ell$ is the loss function.


\subsubsection{Client Side Update.}
At the beginning of each round \(t\), the server randomly selects a subset \(S_t \subset K\) of clients to participate in the federated training process and subsequently shares the current global model \(\theta_r^g\) to these participating clients. Each client updates its local model by initializing with the global model parameters (\(\theta_k^r = \theta_g^r\)) and then updates its local model by minimizing the local objective function. For local training, we have developed an adaptive objective function that balances local loss with the divergence between local and global models. The extent of this divergence is quantified using the Kullback-Leibler (KL) divergence~\citep{csiszar1975divergence}, which effectively compares the probability distributions of the local model weights \( p^k(w) \) with the global model weights \( q(w) \). The KL divergence is mathematically defined in Eq.~\ref{eq2xd}. To obtain the probability distributions of the local and global model weights, we first flatten the weights and then apply the softmax function. This process yields the desired probability distributions (p), as specified in Eq.~\ref{ss}.


\begin{equation}
\label{ss}
\text{p} = \frac{\exp(\text{flatten weights})}{\sum \exp(\text{flatten weights})}
\end{equation}






\begin{equation}
\label{eq2xd}
D_{\text{KL}}(p^k \| q) = \sum_{i} p^k_i(w) \log\left(\frac{p^k_i(w)}{q_i(w)}\right)
\end{equation}
where $p^k_i$ and $q_i$ are the probabilities associated with the $i^{th}$ component of the weight vectors. The local model must excel on local data while maintaining alignment with the global model to enhance overall generalization. This balance between minimizing local loss and aligning with the global model is defined as local adaptive function $\tilde{f}_k(\theta)$ in Eq.~\ref{eq3ry}.

\begin{equation}
\label{eq3ry}
\tilde{f}_k(\theta) = (1-\beta)*f_k(\theta) + \beta *D_{\text{KL}}(p^k \| q),
\end{equation}

where $f_k(\theta)$ is cross-entropy loss for $k^{th}$ client and $\beta$ is a regularization parameter and should be adaptive to account for the performance discrepancy between the local and global models. When the local model substantially outperforms the global model, $\beta$ should increase to enforce greater alignment. Conversely, if the models perform similarly, $\beta$ should decrease, allowing the local model to focus more on local optimization. The definition of $\beta$ is given in Eq.~\ref{eq4oh}.

\begin{equation}
\label{eq4oh}
\beta = \sigma(\mathcal{A}_{\text{local}}^k - \mathcal{A}_{\text{global}}^k)
\end{equation}
where $\sigma$ is the sigmoid function, $\mathcal{A}_{\text{local}}^k$ represents the local model accuracy, and $\mathcal{A}_{\text{global}}^k$ is the global model accuracy for client $k$. We calculated the global model's accuracy $\mathcal{A}_{\text{global}}^k$ for client $k$ by evaluating it on the training data of client $k$ prior to performing local updates in the current round. Incorporating the adaptive parameter $\beta$ in Eq.~\ref{eq3ry}, the adaptive loss function for client $k$ is represented in Eq.~\ref{eq5mi}.

\begin{equation}
\begin{aligned}
\label{eq5mi}
\mathcal{L}_{\text{adaptive}}^k = (1-(\sigma(\mathcal{A}_{\text{local}}^k - \mathcal{A}_{\text{global}}^k))*\mathcal{L}_{\text{local}}^k \\+ 
\sigma(\mathcal{A}_{\text{local}}^k - \mathcal{A}_{\text{global}})* D_{\text{KL}}(p^k \| q)
\end{aligned}
\end{equation}

After defining the adaptive loss function for each client, we optimize the local model parameters using  stochastic gradient descent (SGD). The gradient update for the local model weights $w_k$ based on the adaptive loss function is given in Eq.~\ref{eq6rtw}.
\begin{equation}
\label{eq6rtw}
w_k^{t+1} = w_k^t - \eta \nabla_w \mathcal{L}_{\text{adaptive}}^k(w_k^t),
\end{equation}
where $\eta$ is the local learning rate. Expanding the gradient term $\nabla_w \mathcal{L}_{\text{adaptive}}^k(w_k^t)$, we obtain Eq.~\ref{eq7mq}:

\begin{equation}
\label{eq7mq}
\begin{aligned}
\nabla_w \mathcal{L}_{\text{adaptive}}^k(w_k^t) = (1-(\sigma(\mathcal{A}_{\text{local}}^k - \mathcal{A}_{\text{global}}^k))\nabla_w \mathcal{L}_{\text{local}}^k(w_k^t) + \sigma(\mathcal{A}_{\text{local}}^k - \mathcal{A}_{\text{global}}) \nabla_w D_{\text{KL}}(p^k \| q).
\end{aligned}
\end{equation}

The KL divergence term, \(\sigma(\mathcal{A}_{\text{local}}^k - \mathcal{A}_{\text{global}}^k) \nabla_w D_{\text{KL}}(p^k \| q)\) in Eq.~\ref{eq7mq}, acts as a regularizer to keep the local model gradients aligned with the global model gradients, thereby preserving model coherence despite non-IID data. The adaptive coefficient $\beta$ (Eq.~\ref{eq4oh}) is dynamically computed as a function of the performance gap $(A^k_{\text{local}} - A^k_{\text{global}})$. When the global model outperforms the local one, $\beta$ tends toward 0 (via the sigmoid function), thus increasing the weight on the local loss term $(1 - \beta)$ and enabling the client to focus more on its local data. Conversely, when the local model performs better than the global, $\beta$ increases toward 1, strengthening the KL term to preserve global knowledge rather than blindly aligning the models. 
The rationale is that superior local accuracy often reflects \textit{overfitting} to non-IID client data. A higher $\beta$ in such cases regularizes the local model by constraining it to remain close to the global parameter manifold, improving overall generalization. Conversely, when the global model generalizes better, a smaller $\beta$ allows the client to emphasize local learning. For example, a client trained only on classes 0--4 may achieve high local accuracy but would poorly generalize to unseen classes 5--9. The higher $\beta$ and associated KL regularization preserve the global model’s multi-class knowledge, thereby preventing catastrophic forgetting.
















Note that we employed KL divergence on the softmax of the parameters to capture distributional alignment between local and global models rather than direct numerical differences. By transforming flattened parameters into probability distributions through softmax, we interpret model weights as expressing relative importance rather than absolute magnitude. KL divergence thus quantifies how the structural configuration of model parameters diverges between clients and the server.
Compared to alternatives, logit-based distances capture only output-level similarity and overlook the underlying parameter drift that degrades generalization. Layer-wise representation distances would require computing activations on a reference dataset, introducing computational overhead and ambiguity regarding data selection in heterogeneous settings. Similarly, the L2 parameter distance (as in FedProx) reflects magnitude differences but not distributional structure which is an essential factor for understanding model behavior. Empirically, we find KL divergence advantageous because its asymmetry aligns with the local to global adaptation objective and its bounded nature ensures stable gradients during optimization.

\subsubsection{Server Side Update.}
After obtaining the weights from the participating clients at round $t$, the server calculates the Wasserstein Barycenter to effectively aggregate the weights of the last layers of the client models. Computing exact Wasserstein Barycenter can be computationally expensive, so we have approximated it using the Sinkhorn-Knopp~\citep{
knight2008sinkhorn} algorithm for efficient computation. We consider the local model weights as distributions and assign equal importance to each client in the computation of the Wasserstein Barycenter ($\bar{\mu}$). This barycenter represents the distribution that minimizes the sum of Wasserstein distances to the individual client gradient distributions, as formally defined in Eq.~\ref{eq8}.

\begin{equation}
\label{eq8}
    \hat{\mu} = \arg\min_{\nu} \sum_{k=1}^K \lambda_k W(\mu_k, \nu)
\end{equation}

where $\lambda_k$ are weights corresponding to the importance or reliability of the client $k$. The Wasserstein distance $W(\mu_k, \mu_j)$ between two gradient distributions $\mu_k$ and $\mu_j$ of clients $j$ and $k$ is defined in Eq.~\ref{eq9}.

\begin{equation}
\label{eq9}
W(\mu_k, \mu_j) = \left( \inf_{\gamma \in \Gamma(\mu_k, \mu_j)} \int_{\mathcal{X} \times \mathcal{X}} d(x, y)^p \, d\gamma(x, y) \right)^{1/p}
\end{equation}

where $\Gamma(\mu_k, \mu_j)$ denotes the set of all couplings (or joint distributions) $\gamma$ on $\mathcal{X} \times \mathcal{X}$ with marginals $\mu_k$ and $\mu_j$ respectively, and $d(x, y)$ is the distance between points $x$ and $y$ in the metric space $\mathcal{X}$. After that, we use Sinkhorn-Knopp algorithm to calculate the Wasserstein Barycenter. 



This barycenter is computed iteratively, starting by calculating a scaling factor \(\gamma\) using Eq.~\ref{eq11}, followed by Eq.~\ref{aa}.


\begin{equation}
\label{eq11}
\gamma = \exp\left(-\frac{\text{W}(\bar{p}, p_i)}{\epsilon}\right)
\end{equation}

\begin{equation}
\label{aa}
\bar{p}_{\text{new}} = \frac{\sum_i \lambda_i \gamma p_i}{\sum_i \lambda_i \gamma_i}
\end{equation}

where \( \bar{p} \) is the current estimate of the barycenter, $p_i$ refers to the $i^{th}$  client's gradient distribution, \( \epsilon \) is a small positive constant, and the iterations continue until convergence. After few iterations, we get the Wasserstein barycenter that is used to update the global model weights. We update the the global model weights for the last layers by substracting them from the calculated Wasserstein barycenter for effectively aggregating the updates from the last layers. Our gradient analysis (Fig.~\ref{fig:grad_norm_comparison}) indicates that non-IID data disproportionately amplifies gradient norms in the final layers, leading to heightened instability compared to IID conditions. Traditional averaging methods aggregate client updates in Euclidean space without accounting for the underlying distributional differences across clients. The Wasserstein Barycenter addresses this limitation by operating in the space of probability distributions, finding the optimal aggregation point that minimizes distributional discrepancies (Wasserstein distances) to all client updates. This geometry-aware approach provides more robust aggregation by explicitly accounting for heterogeneous client learning behaviors, reducing the bias and instability caused by non-IID data. The algorithm of proposed method FedDUAL is given in the Algorithm ~\ref{alg:fed_wasserstein}. The proof of the convergence for both convex and non-convex settings for the proposed method can be found in Section~\ref{tgyhu} of the Appendix.

\begin{algorithm}[h]
\caption{FedDUAL}
\label{alg:fed_wasserstein}
\begin{algorithmic}[1]
\item {\textbf{Input:}} Number of clients $K$, Number of communication rounds $T$, and Global model $\mathcal{G}$.
\item {\textbf{Output:}} Trained global model $\mathcal{G}^*$.
\State Define a mask $e \in \{0, 1\}^d$, where $e_j = 1$ for the last few layers and $0$ for the rest layers.
\State Let $S_{naive} = \{j : e_j = 0\}$ and $S_{dynamic} = \{j : e_j = 1\}$.
\State \textbf{Initialize} global model weights $\theta^g$ 
\For{$t = 1$ to $T$}
    \State Sample a subset of clients $\mathcal{S}_t \subseteq \{1, \ldots, K\}$
    \State Initialize lists: local model weights $\mathcal{W} \gets []$, gradients $\Delta \gets []$ 
    \For{each client $k \in \mathcal{S}_t$}
        \State Initialize local model $\mathcal{M}_k$ with global weights $\theta^g$.
        \State Train $\mathcal{M}_k$ on local dataset $\mathcal{D}_k$ using adaptive loss function defined in Eq.~\ref{eq5mi}.
        \State $\mathcal{W} \gets \mathcal{W} \cup \{\theta_k\}$ \Comment{Store local model weights $\theta_k$}
        \State Compute gradients $\nabla_k$ for $\mathcal{M}_k$
        \State $\Delta \gets \Delta \cup \{\nabla_k\}$ \Comment{Store gradients $\nabla_k$}
    \EndFor
    
    \For{$j \in \{1, \ldots, d\}$}
        \If{$e_j = 1$}  \Comment{Layer belongs to $S_{dynamic}$}
            \State Extract last layers' gradients $\{\nabla_k[j]\}$ from $\Delta$
            \State Compute Wasserstein Barycenter of last layer $j$ gradients $\bar{\nabla}_j$
            \State Update global model's last layer $j$ weights $\theta^g[j] \gets \theta^g[j] - \bar{\nabla}_j$
        \Else \Comment{Layer belongs to $S_{naive}$}
            \State Perform Federated Averaging for layer $j$: 
            \State $\theta^g[j] \gets \frac{1}{|\mathcal{S}_t|} \sum_{k \in \mathcal{S}_t} \theta_k[j]$
        \EndIf
    \EndFor
    
\EndFor
\State $\mathcal{G}^* \gets \theta^g$ \Comment{Final trained global model}
\end{algorithmic}
\end{algorithm}

\begin{table}[ht]
\caption{Top-1 accuracy (\%) on CIFAR10, CIFAR100, and FMNIST datasets. The values in bold represent the highest accuracy achieved. '*' denotes algorithms that failed to achieve convergence.}
\centering
\scalebox{1}{
\begin{tabular}{lccc}
\toprule
& CIFAR10 & CIFAR100 & FMNIST \\
\midrule
FedAvg & 46.68 $\pm$ 0.25 & 26.88 $\pm$ 0.18 & 81.70 $\pm$ 0.20 \\
\midrule
FedProx & 47.58 $\pm$ 0.30 & 26.89 $\pm$ 0.22 & 80.54 $\pm$ 0.28 \\
FedNova & 48.44 $\pm$ 0.35 & * & * \\
FedBN & * & 26.88 $\pm$ 0.19 & 81.36 $\pm$ 0.23 \\
FedDyn & 43.97 $\pm$ 0.40 & 18.27 $\pm$ 0.32 & 71.86 $\pm$ 0.45 \\
MOON & 46.57 $\pm$ 0.28 & 28.50 $\pm$ 0.25 & 80.09 $\pm$ 0.27 \\
SCAFFOLD & * & * & * \\
FedPVR & 42.26 $\pm$ 0.42 & 23.78 $\pm$ 0.31 & 80.32 $\pm$ 0.33 \\
\midrule
\textbf{Proposed} & \textbf{48.70 $\pm$ 0.20} & \textbf{29.15 $\pm$ 0.24 }& \textbf{81.99 $\pm$ 0.21 }\\
\bottomrule
\end{tabular}}
\label{result1}
\end{table}

\begin{figure*}
\centering
\includegraphics[width=\textwidth]{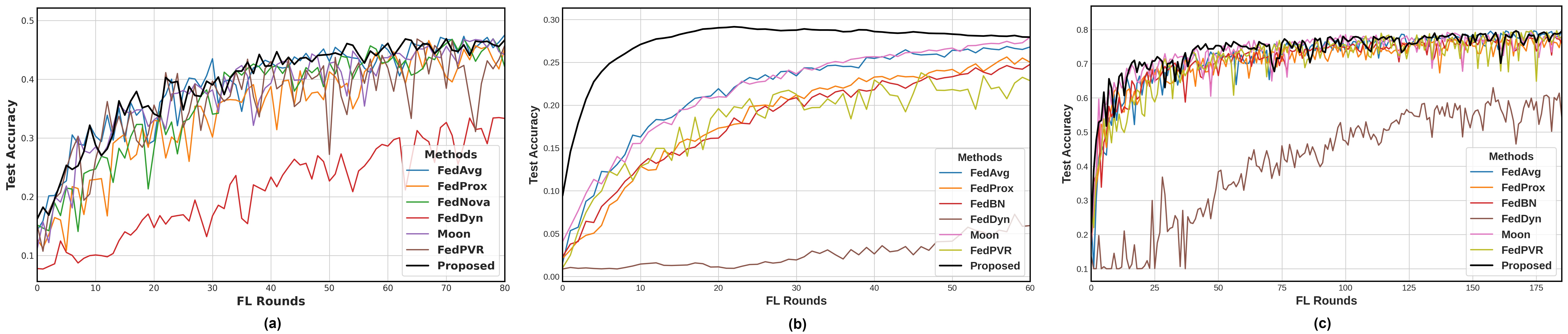}
\caption{ Learning curves comparing the proposed method with baselines across various datasets: (a) CIFAR-10, (b) CIFAR-100, and (c) FMNIST.}
\label{fig3}
\end{figure*}

\begin{figure*}
\centering
\includegraphics[width=\linewidth]{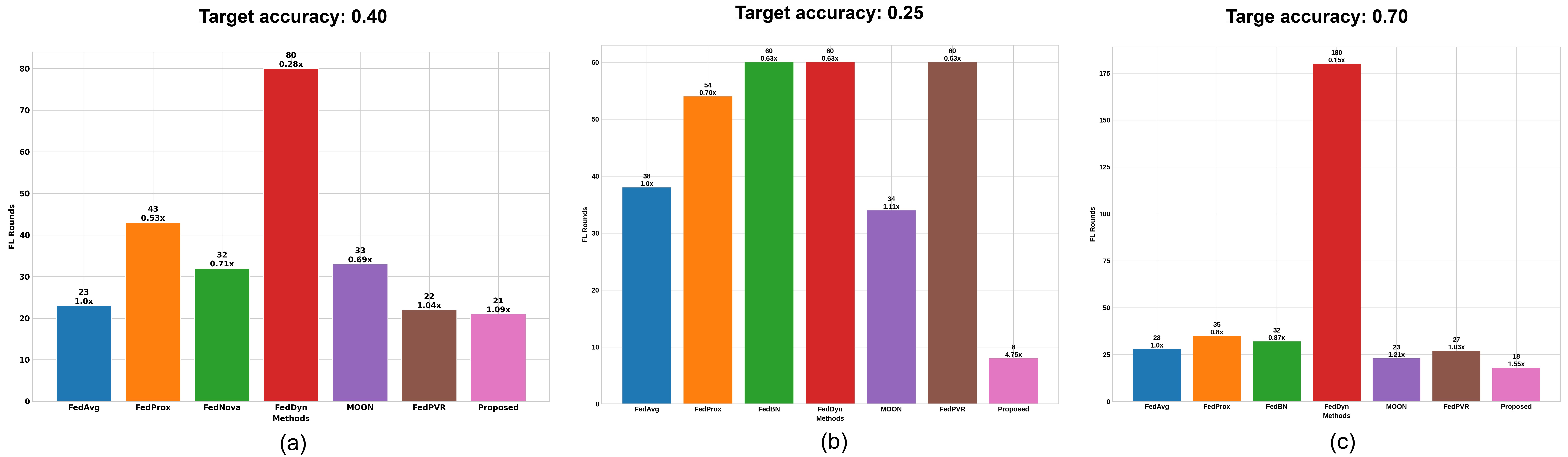}
\caption{Number of FL rounds required to reach the target accuracy for the proposed method and other baselines on different datasets: (a) CIFAR-10, (b) CIFAR-100, and (c) FMNIST.}
\label{fig4}
\end{figure*}

\section{Experimental Results}
\subsection{Experimental Setup}

To assess the effectiveness of the proposed FedDUAL approach, we conducted extensive experiments using three widely recognized classification benchmarks: CIFAR10~\citep{krizhevsky2009learning}, CIFAR100~\citep{Krizhevsky09learningmultiple}, and FMNIST~\citep{xiao2017fashion}. To simulate real-world non-IID data distributions, we employed a client-wise partitioning strategy based on the Dirichlet distribution~\citep{hsu2019measuring}. This distribution is governed by a concentration parameter $\alpha$, which controls the degree of data heterogeneity among clients. Lower $\alpha$ values result in more skewed data distributions, closely mimicking uneven data partitions. In all experiments, we set $\alpha$ = 0.01 to simulate severe data heterogeneity, closely approximating real-world conditions. Throughout the communication rounds, each client retains a fixed local data partition. To evaluate the global model's classification performance, we use a separate test dataset maintained at the server, which remains unseen during training. For our experiments, we used LeNet~\citep{lecun1998gradient} for FMNIST dataset and a pre-trained VGG16~\citep{simonyan2015deepconvolutionalnetworkslargescale} for CIFAR-10 and CIFAR-100 dataset, following the methodology outlined in~\citep{hu2024fedmut}. We applied the proposed dynamic aggregation mechanism only to the last two layers of these models. Our setup involved $100$ clients, with 10\% randomly sampled per communication round, and a batch size of $32$. Each client performed three local epochs of model updates. We have computed each result three times with different seed values and reported the mean value with standard deviation. To determine the optimal client learning rate for each experiment, we conducted a grid search over ${0.05, 0.01, 0.2, 0.3}$. For the baseline FedProx, we tested proximal values of {0.001, 0.1, 0.4, 0.7} to find the optimal setting, and for FedNova, we evaluated proximal SGD values from {0.001, 0.003, 0.05, 0.1}, following the recommendations in~\cite{Li_2024}. Across all experiments, we used the Adam optimizer for consistency. We have run each algorithm three times and reported the average outcome. The experimental setup utilized an NVIDIA Quadro RTX 4000 GPU boasting 40GB of memory. The implementation was crafted using Python~\footnote{\url{https://www.python.org/}}, leveraging the TensorFlow framework~\footnote{\url{https://www.tensorflow.org/}} utilizing Windows 11.


\subsection{Comparison with the State-of-the-art Methods}
\subsubsection{Baseline.}
We evaluate the proposed FedDUAL method against eight notable state-of-the-art (SOTA) FL baselines, including FedAvg~\citep{mcmahan2017communication}, FedProx~\citep{li2020federated}, FedNova~\citep{wang2020tackling}, SCAFFOLD~\citep{karimireddy2020scaffold}, FedBN~\citep{li2021fedbn}, FedDyn~\citep{acar2021federated}, MOON~\citep{li2021model} and FedPVR~\citep{li2023effectiveness}. 

\subsubsection{Comparison of Accuracy.}
The results, detailed in Table~\ref{result1}, reveal that many recent FL methods often fall short compared to the standard FedAvg baseline. In contrast, our proposed method consistently achieves SOTA performance, surpassing FedAvg along with other baselines across all evaluated scenarios. Furthermore, our approach exhibits remarkable adaptability across diverse datasets. Unlike some algorithms that excel on specific datasets but falter on others, the proposed FedDUAL consistently outperforms baselines across a wide range of data environments. This improvement suggests that our method addresses fundamental challenges in FL, potentially offering a more generalizable solution to the issues posed by data heterogeneity in federated settings. We also observed that FedNova, FedBN, and Scaffold did not perform effectively in our experimental setup.

\subsubsection{Comparison of Convergence.}
Figure~\ref{fig3} compares the learning curves of our method with baselines, while Fig.~\ref{13jhjh01} in the Appendix includes the corresponding curves with error bars. Across all datasets, our method consistently converges faster and achieves higher final accuracy. Although the number of communication rounds varies by dataset, performance generally saturates by the final round. Notably, our method not only attains a more robust final model but also displays markedly faster convergence across all datasets examined. This effectiveness is further highlighted in Fig.~\ref{fig4}, where it consistently reaches target accuracy with far fewer communication rounds compared to baseline approaches. However, the proposed method incurs a higher computation cost due to Wasserstein barycenter based aggregation at the server side. We have provided a thorough computational complexity analysis and runtime analysis in the Section~\ref{xdbb} and the Section~\ref{run} of the appendix respectively.


\subsection{Validation of the Motivation}



To substantiate our claim that the proposed method yields models in flatter loss landscapes compared to FedAvg, we conducted a comparative analysis. Using VGG-16 models trained on the FMNIST dataset under non-IID conditions ($\alpha=0.01$), we visualized their respective loss landscapes following the approach outlined in~\cite{li2018visualizing}. Figure~\ref{fig6} in the Appendix depicts these landscapes, with each model centrally located within its respective terrain. The visualization reveals that our proposed method situates the model in a notably flatter region compared to FedAvg. This finding supports our assertion that our approach guides federated training towards more stable and generalizable solutions, characterized by flatter loss landscapes. The performance improvement of the proposed models stems from two key innovations: a Wasserstein Barycenter-based aggregation for final layer gradients, mitigating client drift in heterogeneous data environments, and an adaptive loss function balancing local optimization with global consistency during client training. This synergistic approach preserves global knowledge while promoting client-specific optimization, addressing fundamental FL challenges.

\section{Ablation Study}

In our ablation study, we performed all experiments on the FMNIST dataset with $\alpha = 0.01$. The study comprised four types of experiments: (1) performance analysis of the individual modules, (2) assessment of the impact of dynamic aggregation across different neural network layers, (3) hyperparameter analysis, and (4) evaluation of various levels of data heterogeneity. 

\subsubsection{Performance Analysis of Individual Modules}

To assess the effectiveness of the proposed adaptive loss and dynamic aggregation techniques, we conducted three ablation experiments across FMNIST, CIFAR-10, and CIFAR-100. The results for FMNIST are shown in Table~\ref{tab:ablation_combined}. In the first experiment, we employed only the adaptive loss alongside standard server-side aggregation. Notably, this configuration underperforms FedAvg across all datasets, indicating that adaptive loss alone cannot effectively address data heterogeneity—likely due to its limited capacity to improve generalization despite fostering local-global alignment. The second experiment implemented our dynamic aggregation technique at the server, while retaining the conventional cross-entropy loss function locally. Finally, the third experiment combined both proposed methods: the adaptive loss function and the dynamic aggregation technique. As evidenced by Table~\ref{tab:ablation_combined}, the integration of both proposed approaches in the third experiment yielded the highest accuracy, highlighting the impact of our dual strategy on model performance. The learning curves for these experiments using FMNIST dataset are illustrated in Fig.~\ref{fig11} of the Appendix.




\begin{wraptable}{r}{0.5\textwidth}
\vspace{-10pt}
\caption{Ablation study of FedDUAL across different datasets.}
\centering
\scriptsize
\setlength{\tabcolsep}{3pt}
\begin{tabular}{@{}c c c c@{}}
\hline
Adaptive Loss & Dynamic Agg. & Dataset & Acc. (\%) \\  
\hline
\ding{51} & \ding{55} & \multirow{3}{*}{FMNIST} & 80.70 $\pm$ 0.22 \\  
\ding{55} & \ding{51} &  & 80.91 $\pm$ 0.25 \\  
\ding{51} & \ding{51} &  & \textbf{81.99 $\pm$ 0.18} \\  
\hline
\ding{51} & \ding{55} & \multirow{3}{*}{CIFAR10} & 41.05 $\pm$ 0.12 \\  
\ding{55} & \ding{51} &  & 46.50 $\pm$ 0.15 \\  
\ding{51} & \ding{51} &  & \textbf{48.70 $\pm$ 0.20} \\  
\hline
\ding{51} & \ding{55} & \multirow{3}{*}{CIFAR100} & 25.05 $\pm$ 0.11 \\  
\ding{55} & \ding{51} &  & 27.01 $\pm$ 0.17 \\  
\ding{51} & \ding{51} &  & \textbf{29.15 $\pm$ 0.24} \\  
\hline
\end{tabular}
\label{tab:ablation_combined}
\end{wraptable}

\subsubsection{Impact of Dynamic Aggregation on Different Network Layers}
\begin{wrapfigure}{r}{0.5\textwidth} 
    \centering
    \includegraphics[width=0.44\textwidth]{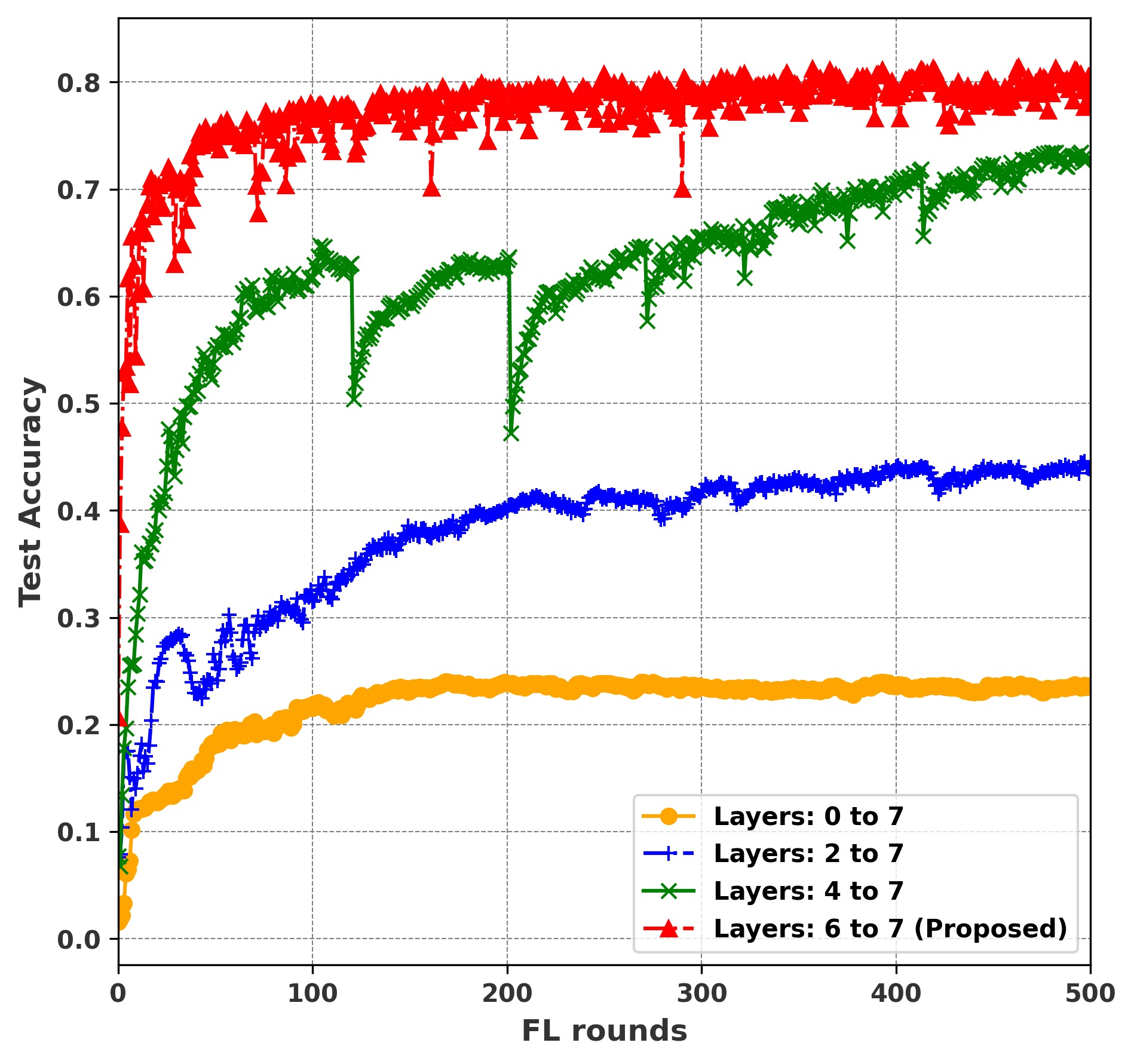} 
    \caption{Illustration of the Dynamic aggregation method applied across various layers of the neural network.}
    \label{fig7}
\end{wrapfigure}

To substantiate our decision to apply dynamic aggregation technique selectively to last layers, we examined its impact across various layers of the neural network. Our earlier findings highlighted that data heterogeneity primarily affects last layers of the network. Figure~\ref{fig7} illustrates that random utilization of the dynamic aggregation to all layers diminishes performance. Conversely, targeted implementation on layers proximal to the classifier yielded optimal accuracy and convergence. These outcomes validate our hypothesis and demonstrate the method's efficacy in mitigating heterogeneity-induced issues. By focusing our dynamic aggregation technique on the most susceptible layers, we directly address the core challenge of data heterogeneity in federated training, resulting in enhanced model performance and faster convergence.

\subsubsection{Hyperparameter Analysis}
In the proposed architecture, there are two key hyperparameters to consider: the scaling factor ($\gamma$) and the number of iterations used to compute the Wasserstein Barycenter. The proposed FedDUAL approach utilizes dynamic server-side aggregation by applying the Wasserstein Barycenter concept to combine the weights of the final layers from local models. This iterative process involves a small positive constant ($\epsilon$) to determine the scaling factor ($\gamma$). To optimize performance, we conducted two sets of experiments on the FMNIST dataset with $\alpha=0.01$, each exploring a range of values for these crucial hyperparameters. The hyperparameter $\epsilon$ influences the sensitivity of the barycenter calculation to variations in Wasserstein distance. A smaller $\epsilon$ makes the barycenter more responsive to differences in Wasserstein distance, while a larger $\epsilon$ diminishes this sensitivity. This impacts how the barycenter integrates each distribution according to its distance from the current estimate. During the iterative update of the barycenter, $\epsilon$ affects the scaling factor $\gamma$ applied to each distribution. An excessively small $\epsilon$ can result in slow or potentially non-existent convergence due to minimal scaling factor, whereas a too-large $\epsilon$ may cause oversmoothing, reducing the barycenter's effectiveness in accurately representing the distributions. For this setting we have fixed the number of iterations to compute Wasserstein Barycenter as 150. Figure~\ref{fig9} in the Appendix shows test accuracy across different $\epsilon$ values, indicating that larger $\epsilon$ can degrade performance or hinder convergence. Figure~\ref{fig10} in the Appendix presents the corresponding learning curves for these settings. The number of iterations in the Wasserstein Barycenter function is another critical hyperparameter that affects both the accuracy and efficiency of the barycenter computation. Generally, more iterations enhance convergence and accuracy, ensuring that the barycenter more closely approximates the optimal value. However, increasing the number of iterations also prolongs computation time, necessitating a balance between accuracy and efficiency. Finding the optimal number of iterations involves a trade-off: too few iterations may result in suboptimal outcomes, while too many can yield diminishing returns in accuracy. To achieve the best performance, begin with a reasonable default value, monitor convergence by observing changes in the barycenter, and adjust iteratively based on empirical results and available computational resources. For this setting, we fixed the $epsilon$ value as 0.0001, which yields the highest results in previous experiment. Figure~\ref{fig12} illustrates the test accuracy for different values of iterations to calculate Wasserstein Barycenter, suggesting that larger iterations may adversely affect performance. Figure~\ref{fig13} presents the corresponding learning curves for these settings. From both experiments, we observe that the highest performance is achieved with $\epsilon = 0.00001$ and 150 iterations. Therefore, to optimize performance, it is advisable to set $\epsilon$ to a smaller value while keeping the number of iterations between 100 and 150.

\subsubsection{Experiment on Different Level of Data Heterogeneity}
Figure~\ref{fig131} illustrates the accuracy of the proposed method and various baselines across different levels of data heterogeneity on the FMNIST dataset. In this context, heterogeneity is quantified by $\alpha$, with lower values indicating greater data heterogeneity. The results show that as $\alpha$ decreases, the test accuracy for all models increases, because data heterogeneity among clients is decreased. Remarkably, the proposed method consistently achieves the highest test accuracy and exhibits the slowest performance decline compared to other algorithms, demonstrating superior performance of the proposed method on varying degrees of non-IID data partitioning. The learning curve is presented in Fig.~\ref{1301gyg}.

\section{Conclusion}
This research presents a novel approach to address the challenges posed by data heterogeneity among clients in the federated approach. We systematically analyze the factors contributing to federated model performance degradation under severe data heterogeneity and propose an architecture incorporating dual-strategy innovations. First, we implement an adaptive loss function for client-side training. Second, we create a dynamic aggregation strategy for server side aggregation, tailored to client-specific learning behaviors. The proposed FedDUAL effectively overcomes the challenges of heterogeneous data, outperforming eight SOTA baselines. It demonstrates faster convergence and consistently improved performance, making it an excellent solution for large-scale FL applications in real-world scenarios. Our approach's flexibility paves the way for research into hybrid federated learning models that adapt to changing client environments and data. Future studies will focus on integrating personalized learning paths to enhance model adaptability and efficiency across various datasets.

\bibliographystyle{tmlr}
\bibliography{tmlr}

\appendix
\section*{Appendix}

\section{Convergence Proof}
\label{tgyhu}

Before presenting the main convergence theorems, we establish assumptions and several key lemmas.
\subsection*{Assumptions}
\textbf{1. L-Smoothness:}
Each local loss function $f_k(\theta)$ is $L$-smooth:
\begin{equation}
\lVert\nabla f_k(\theta) - \nabla f_k(\theta')\rVert \leq L \lVert\theta - \theta'\rVert, \quad \forall \theta, \theta' \in \mathcal{W}, \forall k
\end{equation}

\textbf{2. Unbiased Stochastic Gradients:}
For any client $k$ and parameter $\theta$, the stochastic gradient is unbiased:
\begin{equation}
\E_{\xi \sim \mathcal{D}_k}[\nabla \ell(\theta; \xi)] = \nabla f_k(\theta).
\end{equation}

\textbf{3. Bounded Variance:}
The variance of stochastic gradients is bounded:
\begin{equation}
\mathbb{E}_{\xi \sim \mathcal{D}_k}\left[\lVert\nabla \ell(\theta; \xi) - \nabla f_k(\theta)\rVert^2\right] \leq \sigma^2, \quad \forall \theta, k
\end{equation}
\textbf{where $\sigma^2 = \max_k \sigma_k^2$.}

\textbf{4. Bounded Gradients:}
There exists $G > 0$ such that:
\begin{equation}
\lVert\nabla f_k(\theta)\rVert \leq G, \quad \forall \theta \in \mathcal{W}, \forall k
\end{equation}

\textbf{5. KL Divergence Properties:}
The KL divergence term and dynamic weighting satisfy:
\begin{equation}
\lVert\nabla D_{\text{KL}}(p_k(\theta) \| q(\theta))\rVert \leq G_{\text{KL}}, \quad \forall \theta, k
\end{equation}
\begin{equation}
\lVert\nabla \beta_k(\theta)\rVert \leq L_{\beta}, \quad \forall \theta, k
\end{equation}
\begin{equation}
0 \leq \beta_k(\theta) \leq \beta_{\max} < 1, \quad \forall \theta, k
\end{equation}

\textbf{6. Wasserstein Barycenter Approximation:}
For the final layers aggregated using Wasserstein Barycenters, the approximation error $\varepsilon_{\text{WB}}$ is bounded as follows:
\begin{equation}
\lVert\nabla^{\text{WB}}_t - \nabla^{\text{exact}}_t\rVert \leq \varepsilon_{\text{WB}}
\end{equation}
where $\nabla^{\text{exact}}_t$ represents the exact gradient aggregation.

\subsection*{A.1 Key Lemmas:}

\textbf{Lemma 1: Smoothness of Modified Loss function:}
\label{lemma:smooth_modified}\\

Under Assumptions 1 and 5, the modified loss function $\tilde{f}_k(\theta)$ is $\tilde{L}$-smooth with
\begin{equation}
\label{eq:smoothness_constant}
\tilde{L} = L + L_{\beta}(G + G_{\text{KL}}) + \beta_{\max} G_{\text{KL}} L_{\beta} + L_{\beta} G_f,
\end{equation}

where $G_f = \max_k \sup_\theta \|f_k(\theta)\|$ and $G_{\text{KL}} = \max_k \sup_\theta \|D_{\text{KL}}(p_k(\theta) \| q(\theta))\|$.

\textbf{Proof.}  
Let $\theta, \theta' \in \mathcal{W}$. The complete gradient of the modified loss function is:

\begin{equation}
\begin{aligned}
\nabla \tilde{f}k(\theta) &= \frac{\partial}{\partial \theta}\left[(1 - \beta_k(\theta)) f_k(\theta) + \beta_k(\theta) D{\text{KL}}(p_k(\theta) | q(\theta))\right] 
\\
&= -\nabla \beta_k(\theta) \cdot f_k(\theta) + (1 - \beta_k(\theta)) \nabla f_k(\theta) \
\\
&\quad + \nabla \beta_k(\theta) \cdot D_{\text{KL}}(p_k(\theta) | q(\theta)) + \beta_k(\theta) \nabla D_{\text{KL}}(p_k(\theta) | q(\theta))
\end{aligned}
\end{equation}

Similarly for $\theta'$:
\begin{align}
\nabla \tilde{f}_k(\theta') &= -\nabla \beta_k(\theta') \cdot f_k(\theta') + (1 - \beta_k(\theta')) \nabla f_k(\theta') \nonumber \\
&\quad + \nabla \beta_k(\theta') \cdot D_{\text{KL}}(p_k(\theta') \| q(\theta')) + \beta_k(\theta') \nabla D_{\text{KL}}(p_k(\theta') \| q(\theta'))
\end{align}

Calculating the gradient difference results in Eq.~\ref{o0o0}:

\begin{equation}
\begin{aligned}
\nabla \tilde{f}k(\theta) - \nabla \tilde{f}k(\theta')
\\
&= -\left[\nabla \beta_k(\theta) \cdot f_k(\theta) - \nabla \beta_k(\theta') \cdot f_k(\theta')\right] \
\\
&\quad + \left[(1 - \beta_k(\theta)) \nabla f_k(\theta) - (1 - \beta_k(\theta')) \nabla f_k(\theta')\right] \
\\
&\quad + \left[\nabla \beta_k(\theta) \cdot D_{\text{KL}}(p_k(\theta) | q(\theta)) - \nabla \beta_k(\theta') \cdot D_{\text{KL}}(p_k(\theta') | q(\theta'))\right] \
\\
&\quad + \left[\beta_k(\theta) \nabla D_{\text{KL}}(p_k(\theta) | q(\theta)) - \beta_k(\theta') \nabla D_{\text{KL}}(p_k(\theta') | q(\theta'))\right].
\label{o0o0}
\end{aligned}
\end{equation}

We bound each term separately using the triangle inequality:

\textbf{Term 1:} $-[\nabla \beta_k(\theta) \cdot f_k(\theta) - \nabla \beta_k(\theta') \cdot f_k(\theta')]$
\\

Adding and subtracting $\nabla \beta_k(\theta) \cdot f_k(\theta')$:

\begin{equation}
\begin{aligned}
&\|\nabla \beta_k(\theta) \cdot f_k(\theta) - \nabla \beta_k(\theta') \cdot f_k(\theta')\| \\
&\leq \|\nabla \beta_k(\theta) \cdot (f_k(\theta) - f_k(\theta'))\| + \|(\nabla \beta_k(\theta) - \nabla \beta_k(\theta')) \cdot f_k(\theta')\|  \\
&\leq \|\nabla \beta_k(\theta)\| \cdot \|f_k(\theta) - f_k(\theta')\| + \|\nabla \beta_k(\theta) - \nabla \beta_k(\theta')\| \cdot \|f_k(\theta')\| \\
&\leq L_{\beta} \cdot L \|\theta - \theta'\| + L_{\beta} \|\theta - \theta'\| \cdot G_f  \\
&= L_{\beta}(L + G_f) \|\theta - \theta'\|
\end{aligned}
\end{equation}


\textbf{Term 2:} $[(1 - \beta_k(\theta)) \nabla f_k(\theta) - (1 - \beta_k(\theta')) \nabla f_k(\theta')]$
\\

Adding and subtracting $(1 - \beta_k(\theta)) \nabla f_k(\theta')$:
\begin{equation}
\begin{aligned}
&\|(1 - \beta_k(\theta)) \nabla f_k(\theta) - (1 - \beta_k(\theta')) \nabla f_k(\theta')\| \\
&\leq \|(1 - \beta_k(\theta))(\nabla f_k(\theta) - \nabla f_k(\theta'))\| + \|(\beta_k(\theta') - \beta_k(\theta)) \nabla f_k(\theta')\|  \\
&\leq (1 - \beta_k(\theta)) L \|\theta - \theta'\| + |\beta_k(\theta') - \beta_k(\theta)| \cdot G  \\
&\leq L \|\theta - \theta'\| + L_{\beta} \|\theta - \theta'\| \cdot G  \\
&= (L + L_{\beta} G) \|\theta - \theta'\|
\end{aligned}
\end{equation}

\textbf{Term 3:} $[\nabla \beta_k(\theta) \cdot D_{\text{KL}}(p_k(\theta) \| q(\theta)) - \nabla \beta_k(\theta') \cdot D_{\text{KL}}(p_k(\theta') \| q(\theta'))]$
\\

Following similar decomposition:
\begin{equation}
\begin{aligned}
&\|\nabla \beta_k(\theta) \cdot D_{\text{KL}}(p_k(\theta) \| q(\theta)) - \nabla \beta_k(\theta') \cdot D_{\text{KL}}(p_k(\theta') \| q(\theta'))\| \\
&\leq L_{\beta} G_{\text{KL}} L_{\beta} \|\theta - \theta'\| + L_{\beta} G_{\text{KL}} \|\theta - \theta'\| \\
&= L_{\beta} G_{\text{KL}}(L_{\beta} + 1) \|\theta - \theta'\|
\end{aligned}
\end{equation}

\textbf{Term 4:} $[\beta_k(\theta) \nabla D_{\text{KL}}(p_k(\theta) \| q(\theta)) - \beta_k(\theta') \nabla D_{\text{KL}}(p_k(\theta') \| q(\theta'))]$

\begin{equation}
\begin{aligned}
&\|\beta_k(\theta) \nabla D_{\text{KL}}(p_k(\theta) \| q(\theta)) - \beta_k(\theta') \nabla D_{\text{KL}}(p_k(\theta') \| q(\theta'))\| \\
&\leq \|\beta_k(\theta)(\nabla D_{\text{KL}}(p_k(\theta) \| q(\theta)) - \nabla D_{\text{KL}}(p_k(\theta') \| q(\theta')))\|  \\
&\quad + \|(\beta_k(\theta) - \beta_k(\theta')) \nabla D_{\text{KL}}(p_k(\theta') \| q(\theta'))\|  \\
&\leq \beta_{\max} G_{\text{KL}} L_{\beta} \|\theta - \theta'\| + L_{\beta} G_{\text{KL}} \|\theta - \theta'\|  \\
&= L_{\beta} G_{\text{KL}}(\beta_{\max} + 1) \|\theta - \theta'\|
\end{aligned}
\end{equation}

\textbf{Final bound:} Combining all terms:
\begin{align}
\|\nabla \tilde{f}_k(\theta) - \nabla \tilde{f}_k(\theta')\| &\leq [L_{\beta}(L + G_f) + (L + L_{\beta} G) \\
&\quad + L_{\beta} G_{\text{KL}}(L_{\beta} + 1) + L_{\beta} G_{\text{KL}}(\beta_{\max} + 1)] \|\theta - \theta'\|. \nonumber
\end{align}

For a conservative and simplified bound, we can write:
\begin{equation}
\tilde{L} = L + L_{\beta}(G + G_{\text{KL}} + G_f) + L_{\beta}^2 G_{\text{KL}} + \beta_{\max} L_{\beta} G_{\text{KL}}
\end{equation}

Or more compactly, assuming $G_f \leq G$ and using conservative bounds:
\begin{equation}
\tilde{L} = L + L_{\beta}(G + G_{\text{KL}}) + \beta_{\max} G_{\text{KL}} L_{\beta} + L_{\beta} G_f
\end{equation}

Therefore, $\tilde{f}_k(\theta)$ is $\tilde{L}$-smooth.

\textbf{Lemma 2: Bounded Variance of Modified Gradients:}
\label{lemma:bounded_variance_modified}
Under Assumptions 2, 3, and 5, the variance of stochastic gradients for the modified loss is bounded:
\begin{equation}
\mathbb{E}\left[\lVert\nabla \tilde{f}_k(\theta; \xi) - \nabla \tilde{f}_k(\theta)\rVert^2\right] \leq \sigma^2
\end{equation}



\textbf{Proof.}  
We begin by analyzing the variance of the stochastic gradient of the modified loss function $\tilde{f}_k(\theta)$. Recall that the per-sample stochastic gradient is defined as:
\[
\nabla \tilde{f}_k(\theta; \xi) = (1 - \beta_k(\theta)) \nabla \ell(\theta; \xi) + \beta_k(\theta) \nabla D_{\mathrm{KL}}(p_k(\theta) \,\|\, q(\theta)),
\]
whereas the full-batch gradient is:
\[
\nabla \tilde{f}_k(\theta) = (1 - \beta_k(\theta)) \nabla f_k(\theta) + \beta_k(\theta) \nabla D_{\mathrm{KL}}(p_k(\theta) \,\|\, q(\theta)).
\]

Subtracting the two, we obtain:
\[
\nabla \tilde{f}_k(\theta; \xi) - \nabla \tilde{f}_k(\theta) = (1 - \beta_k(\theta)) \left( \nabla \ell(\theta; \xi) - \nabla f_k(\theta) \right).
\]
Here, the KL divergence term cancels out since it is deterministic and does not depend on the stochastic sample $\xi$. Taking the squared norm and expectation over the stochasticity of $\xi$, we have:
\[
\begin{aligned}
\mathbb{E}_\xi \left[ \left\| \nabla \tilde{f}_k(\theta; \xi) - \nabla \tilde{f}_k(\theta) \right\|^2 \right]
&= (1 - \beta_k(\theta))^2 \cdot \mathbb{E}_\xi \left[ \left\| \nabla \ell(\theta; \xi) - \nabla f_k(\theta) \right\|^2 \right] \\
&\leq \mathbb{E}_\xi \left[ \left\| \nabla \ell(\theta; \xi) - \nabla f_k(\theta) \right\|^2 \right] \quad \text{(since } (1 - \beta_k(\theta))^2 \leq 1 \text{)} \\
&\leq \sigma_k^2 \leq \sigma^2,
\end{aligned}
\]
where we have used Assumption 3 to upper bound the variance of the stochastic gradients by $\sigma^2$.

\textbf{Lemma 3: Local Update Analysis:}
\label{lemma:local_update}
Let $\theta_t^k$ be the local model on client $k$ after $E$ local updates initialized from the global model $\theta_t$. Then, under Assumptions 1–5, the expected squared deviation from the global model after local training satisfies:
\begin{equation}
\mathbb{E}\left[\lVert\theta_t^k - \theta_t + \eta E \nabla \tilde{f}_k(\theta_t)\rVert^2\right] \leq \frac{\eta^2 E^2 \tilde{L}^2}{2} \sum_{e=0}^{E-1} \mathbb{E}\left[\lVert\theta_t^{k,e} - \theta_t\rVert^2\right] + \eta^2 E \tilde{\sigma}^2,
\end{equation}
where $\theta_t^{k,e}$ denotes the local model on client $k$ after $e$ local steps, $\tilde{L}$ is the smoothness constant of $\tilde{f}_k$ (Lemma 1, and $\tilde{\sigma}^2$ is the bounded variance of the modified stochastic gradient (Lemma 2).

\textbf{Proof.}
We begin by expressing the full local model update as a telescoping sum over $E$ local steps:
\[
\theta_t^k = \theta_t - \eta \sum_{e=0}^{E-1} \nabla \tilde{f}_k(\theta_t^{k,e}; \xi_e),
\]
where $\xi_e$ denotes the data sample used in the $e$-th local step.

Rewriting this update in terms of the true gradient at the initial point $\theta_t$, we add and subtract $\nabla \tilde{f}_k(\theta_t)$:
\[
\theta_t^k - \theta_t = -\eta \sum_{e=0}^{E-1} \nabla \tilde{f}_k(\theta_t^{k,e}; \xi_e) 
= -\eta E \nabla \tilde{f}_k(\theta_t) - \eta \sum_{e=0}^{E-1} \left( \nabla \tilde{f}_k(\theta_t^{k,e}; \xi_e) - \nabla \tilde{f}_k(\theta_t) \right).
\]

Rearranging terms gives:
\[
\theta_t^k - \theta_t + \eta E \nabla \tilde{f}_k(\theta_t) = -\eta \sum_{e=0}^{E-1} \left( \nabla \tilde{f}_k(\theta_t^{k,e}; \xi_e) - \nabla \tilde{f}_k(\theta_t) \right).
\]

Taking the norm squared and expectation:
\[
\mathbb{E}\left[ \left\| \theta_t^k - \theta_t + \eta E \nabla \tilde{f}_k(\theta_t) \right\|^2 \right] 
= \eta^2 \, \mathbb{E}\left[ \left\| \sum_{e=0}^{E-1} \left( \nabla \tilde{f}_k(\theta_t^{k,e}; \xi_e) - \nabla \tilde{f}_k(\theta_t) \right) \right\|^2 \right].
\]

Applying Jensen’s inequality (or the inequality $\| \sum a_e \|^2 \leq E \sum \| a_e \|^2$):
\[
\leq \eta^2 E \sum_{e=0}^{E-1} \mathbb{E}\left[ \left\| \nabla \tilde{f}_k(\theta_t^{k,e}; \xi_e) - \nabla \tilde{f}_k(\theta_t) \right\|^2 \right].
\]

Now decompose the difference inside each term:
\[
\begin{aligned}
\mathbb{E}\left[ \left\| \nabla \tilde{f}_k(\theta_t^{k,e}; \xi_e) - \nabla \tilde{f}_k(\theta_t) \right\|^2 \right]
&\leq 2 \, \mathbb{E}\left[ \left\| \nabla \tilde{f}_k(\theta_t^{k,e}; \xi_e) - \nabla \tilde{f}_k(\theta_t^{k,e}) \right\|^2 \right] \\
&\quad + 2 \, \mathbb{E}\left[ \left\| \nabla \tilde{f}_k(\theta_t^{k,e}) - \nabla \tilde{f}_k(\theta_t) \right\|^2 \right].
\end{aligned}
\]

From Lemma 2, the variance of the stochastic gradient is bounded:
\[
\mathbb{E}\left[ \left\| \nabla \tilde{f}_k(\theta_t^{k,e}; \xi_e) - \nabla \tilde{f}_k(\theta_t^{k,e}) \right\|^2 \right] \leq \tilde{\sigma}^2.
\]

From Lemma 1, the gradient of $\tilde{f}_k$ is $\tilde{L}$-Lipschitz:
\[
\left\| \nabla \tilde{f}_k(\theta_t^{k,e}) - \nabla \tilde{f}_k(\theta_t) \right\|^2 \leq \tilde{L}^2 \left\| \theta_t^{k,e} - \theta_t \right\|^2.
\]

Combining the two bounds:
\[
\mathbb{E}\left[ \left\| \nabla \tilde{f}_k(\theta_t^{k,e}; \xi_e) - \nabla \tilde{f}_k(\theta_t) \right\|^2 \right]
\leq 2 \tilde{\sigma}^2 + 2 \tilde{L}^2 \mathbb{E}\left[ \left\| \theta_t^{k,e} - \theta_t \right\|^2 \right].
\]

Substituting back:
\[
\mathbb{E}\left[ \left\| \theta_t^k - \theta_t + \eta E \nabla \tilde{f}_k(\theta_t) \right\|^2 \right]
\leq \eta^2 E \sum_{e=0}^{E-1} \left( 2 \tilde{\sigma}^2 + 2 \tilde{L}^2 \mathbb{E}\left[ \left\| \theta_t^{k,e} - \theta_t \right\|^2 \right] \right).
\]

Grouping constants:
\[
= 2 \eta^2 E \tilde{\sigma}^2 + 2 \eta^2 \tilde{L}^2 \sum_{e=0}^{E-1} \mathbb{E}\left[ \left\| \theta_t^{k,e} - \theta_t \right\|^2 \right].
\]

Finally, simplifying constants and using $\frac{1}{2}$ factor for future algebraic convenience:
\[
\leq \frac{\eta^2 E^2 \tilde{L}^2}{2} \sum_{e=0}^{E-1} \mathbb{E}\left[ \left\| \theta_t^{k,e} - \theta_t \right\|^2 \right] + \eta^2 E \tilde{\sigma}^2.
\]

\section*{Theorem 1 (Convex Convergence)}
Suppose Assumptions 1--6 hold and $\tilde{F}(\theta)$ is convex. Let $\theta^* = \arg\min_{\theta \in W} \tilde{F}(\theta)$, and set the learning rate as $\eta \leq \frac{1}{4\tilde{L}E}$.

Then, FedDUAL guarantees the following convergence bound:
\[
\mathbb{E}[\tilde{F}(\bar{\theta}_T) - \tilde{F}(\theta^*)] \leq \frac{2\|\theta_0 - \theta^*\|^2}{\eta T} + \frac{\eta \tilde{L}E \tilde{\sigma}^2}{K} + \eta \tilde{L}E^2 G^2 + \varepsilon_{WB}G
\]
where $\bar{\theta}_T = \frac{1}{T}\sum_{t=0}^{T-1} \theta_t$ and $\tilde{\sigma}^2$ is as defined in Lemma 2.

\section{Proof}

Since $\tilde{F}(\theta)$ is $\tilde{L}$-smooth and convex, we can write the fundamental smoothness inequality:

\begin{align}
\label{u87}
\tilde{F}(\theta_{t+1}) \leq \tilde{F}(\theta_t) + \langle \nabla \tilde{F}(\theta_t), \theta_{t+1} - \theta_t \rangle + \frac{\tilde{L}}{2} \|\theta_{t+1} - \theta_t\|^2
\end{align}

This is the standard smoothness inequality. For any $\tilde{L}$-smooth function $f$, we have $f(y) \leq f(x) + \langle \nabla f(x), y-x \rangle + \frac{\tilde{L}}{2}\|y-x\|^2$.




The FedDUAL update consists of two phases:

\textbf{(For early layers):} Standard aggregation
\begin{align}
\theta^{\text{early}}_{t+1} = \theta^{\text{early}}_t - \frac{\eta}{K} \sum_{k=1}^K \sum_{e=0}^{E-1} \nabla \tilde{f}_k(\theta^{k,e}_t; \xi_{k,e})
\end{align}

\textbf{(For final layers):} Wasserstein Barycenter aggregation
\begin{align}
\theta^{\text{final}}_{t+1} = \text{WB}(\{\theta^{k,E}_t[j]\}_{k=1}^K) + \delta_{WB}
\end{align}

where $\|\delta_{WB}\| \leq \varepsilon_{WB}$ is the Wasserstein approximation error.

The total update can be written as
\begin{align}
\label{u89}
\theta_{t+1} - \theta_t = -\frac{\eta}{K} \sum_{k=1}^K \sum_{e=0}^{E-1} \nabla \tilde{f}_k(\theta^{k,e}_t; \xi_{k,e}) + \delta_{\text{total}},
\end{align}

where $\|\delta_{\text{total}}\| \leq \varepsilon_{WB}$. 
By substituting Eq.~\ref{u89} into the inner product term defined in Eq.~\ref{u87}, we derive Eq.~\ref{mm}.


\begin{equation}
\begin{aligned}
&\langle \nabla \tilde{F}(\theta_t), \theta_{t+1} - \theta_t \rangle \\
&= \left\langle \nabla \tilde{F}(\theta_t), -\frac{\eta}{K} \sum_{k=1}^K \sum_{e=0}^{E-1} \nabla \tilde{f}_k(\theta^{k,e}_t; \xi_{k,e}) + \delta_{\text{total}} \right\rangle \\
&= -\frac{\eta}{K} \sum_{k=1}^K \sum_{e=0}^{E-1} \langle \nabla \tilde{F}(\theta_t), \nabla \tilde{f}_k(\theta^{k,e}_t; \xi_{k,e}) \rangle + \langle \nabla \tilde{F}(\theta_t), \delta_{\text{total}} \rangle
\label{mm}
\end{aligned}
\end{equation}

For the error term, we apply the Cauchy–Schwarz inequality to obtain:
\begin{align}
\label{7mq}
|\langle \nabla \tilde{F}(\theta_t), \delta_{\text{total}} \rangle| \leq \|\nabla \tilde{F}(\theta_t)\| \|\delta_{\text{total}}\| \leq G \varepsilon_{WB}.
\end{align}


Here, we use Assumption 4, which states that $|\nabla \tilde{F}(\theta_t)| \leq G$. We add and subtract $\nabla \tilde{f}_k(\theta_t)$ from the first term of Eq.~\ref{mm} and obtain Eq.~\ref{u98}:

\begin{equation}
\begin{aligned}
\label{u98}
&\langle \nabla \tilde{F}(\theta_t), \nabla \tilde{f}_k(\theta^{k,e}_t; \xi_{k,e}) \rangle \\
&= \langle \nabla \tilde{F}(\theta_t), \nabla \tilde{f}_k(\theta_t) \rangle + \langle \nabla \tilde{F}(\theta_t), \nabla \tilde{f}_k(\theta^{k,e}_t; \xi_{k,e}) - \nabla \tilde{f}_k(\theta_t) \rangle
\end{aligned}
\end{equation}

The first term of right hand side of Eq.~\ref{u98} gives us:
\begin{align}
\label{8uop}
\frac{1}{K} \sum_{k=1}^K \langle \nabla \tilde{F}(\theta_t), \nabla \tilde{f}_k(\theta_t) \rangle = \langle \nabla \tilde{F}(\theta_t), \frac{1}{K} \sum_{k=1}^K \nabla \tilde{f}_k(\theta_t) \rangle = \|\nabla \tilde{F}(\theta_t)\|^2
\end{align}

By definition, $\nabla \tilde{F}(\theta_t) = \frac{1}{K} \sum_{k=1}^K \frac{n_k}{n} \nabla \tilde{f}_k(\theta_t)$, and assuming uniform data distribution, this simplifies to $\frac{1}{K} \sum_{k=1}^K \nabla \tilde{f}_k(\theta_t)$.

For the second term, we decompose:
\begin{equation}
\begin{aligned}
&\nabla \tilde{f}_k(\theta^{k,e}_t; \xi_{k,e}) - \nabla \tilde{f}_k(\theta_t) \\
&= [\nabla \tilde{f}_k(\theta^{k,e}_t; \xi_{k,e}) - \nabla \tilde{f}_k(\theta^{k,e}_t)] + [\nabla \tilde{f}_k(\theta^{k,e}_t) - \nabla \tilde{f}_k(\theta_t)].
\end{aligned}
\end{equation}

Taking expectation and applying the Cauchy–Schwarz inequality, we obtain:
\begin{equation}
\begin{aligned}
&\mathbb{E}[|\langle \nabla \tilde{F}(\theta_t), \nabla \tilde{f}_k(\theta^{k,e}_t; \xi_{k,e}) - \nabla \tilde{f}_k(\theta_t) \rangle|] \\
&\leq \mathbb{E}[\|\nabla \tilde{F}(\theta_t)\| \|\nabla \tilde{f}_k(\theta^{k,e}_t; \xi_{k,e}) - \nabla \tilde{f}_k(\theta^{k,e}_t)\|] \\
&\quad + \mathbb{E}[\|\nabla \tilde{F}(\theta_t)\| \|\nabla \tilde{f}_k(\theta^{k,e}_t) - \nabla \tilde{f}_k(\theta_t)\|]
\label{uu}
\end{aligned}
\end{equation}

Using Young's inequality with parameter $\alpha > 0$:
\begin{align}
ab \leq \frac{a^2}{2\alpha} + \frac{\alpha b^2}{2}
\label{loo}
\end{align}

Applying Eq.~\ref{loo} into first term in Eq.~\ref{uu}, we obtain:
\begin{equation}
\begin{aligned}
&\mathbb{E}[\|\nabla \tilde{F}(\theta_t)\| \|\nabla \tilde{f}_k(\theta^{k,e}_t; \xi_{k,e}) - \nabla \tilde{f}_k(\theta^{k,e}_t)\|] \\
&\leq \frac{\mathbb{E}[\|\nabla \tilde{F}(\theta_t)\|^2]}{2\alpha} + \frac{\alpha}{2} \mathbb{E}[\|\nabla \tilde{f}_k(\theta^{k,e}_t; \xi_{k,e}) - \nabla \tilde{f}_k(\theta^{k,e}_t)\|^2] \\
&\leq \frac{\mathbb{E}[\|\nabla \tilde{F}(\theta_t)\|^2]}{2\alpha} + \frac{\alpha \tilde{\sigma}^2}{2}.
\end{aligned}
\end{equation}

Here we have used Lemma 2 which bounds the variance of stochastic gradients by $\tilde{\sigma}^2$.

Similarly, we can write for the second term in Eq.~\ref{uu}:
\begin{equation}
\begin{aligned}
\label{8i}
&\mathbb{E}[\|\nabla \tilde{F}(\theta_t)\| \|\nabla \tilde{f}_k(\theta^{k,e}_t) - \nabla \tilde{f}_k(\theta_t)\|] \\
&\leq \frac{\mathbb{E}[\|\nabla \tilde{F}(\theta_t)\|^2]}{2\alpha} + \frac{\alpha \tilde{L}^2}{2} \mathbb{E}[\|\theta^{k,e}_t - \theta_t\|^2].
\end{aligned}
\end{equation}

We use the $\tilde{L}$-smoothness of $\tilde{f}_k$ from Lemma 1. 
By selectively combining the terms from Eq.~\ref{7mq}, Eq.~\ref{u98}, Eq.~\ref{8uop}, and Eq.~\ref{8i}, and substituting them into Eq.~\ref{mm}, we obtain Eq.~\ref{45}.
The choice of $\alpha = \frac{1}{2}$ in Young's inequality is a standard optimization choice that balances the two terms in the bound. When applying Young's inequality $ab \leq \frac{a^2}{2\alpha} + \frac{\alpha b^2}{2}$, setting $\alpha = \frac{1}{2}$ gives equal weight to both the gradient norm term and the variance terms, which minimizes the overall bound and leads to the subsequent steps.

\begin{equation}
\label{45}
\begin{aligned}
&\mathbb{E}[\langle \nabla \tilde{F}(\theta_t), \theta_{t+1} - \theta_t \rangle] \\
&\leq -\eta \mathbb{E}[\|\nabla \tilde{F}(\theta_t)\|^2] + \frac{\eta}{K} \sum_{k=1}^K \sum_{e=0}^{E-1} \left[ \frac{\mathbb{E}[\|\nabla \tilde{F}(\theta_t)\|^2]}{2\alpha} + \frac{\alpha \tilde{\sigma}^2}{2} + \frac{\alpha \tilde{L}^2}{2} \mathbb{E}[\|\theta^{k,e}_t - \theta_t\|^2] \right] + G\varepsilon_{WB}.
\end{aligned}
\end{equation}

Substituting $\alpha = \frac{1}{2}$, we obtain:
\begin{align}
\label{x23}
&\leq -\eta \mathbb{E}[\|\nabla \tilde{F}(\theta_t)\|^2] + \frac{\eta E}{K} \sum_{k=1}^K \left[ \mathbb{E}[\|\nabla \tilde{F}(\theta_t)\|^2] + \frac{\tilde{\sigma}^2}{4} + \frac{\tilde{L}^2}{4} \sum_{e=0}^{E-1} \mathbb{E}[\|\theta^{k,e}_t - \theta_t\|^2] \right] + G\varepsilon_{WB}.
\end{align}

Simplifying Eq.~\ref{x23}, we obtain Eq.~\ref{fjx}:
\begin{align}
\label{fjx}
&\leq -\frac{\eta}{2} \mathbb{E}[\|\nabla \tilde{F}(\theta_t)\|^2] + \frac{\eta \tilde{L}^2 E}{4K} \sum_{k=1}^K \sum_{e=0}^{E-1} \mathbb{E}[\|\theta^{k,e}_t - \theta_t\|^2] + \frac{\eta E \tilde{\sigma}^2}{4} + G\varepsilon_{WB}.
\end{align}

Again we starting from Eq.~\ref{u89}:

\[
\theta_{t+1} - \theta_t = -\eta \cdot \frac{1}{K} \sum_{k=1}^K \sum_{e=0}^{E-1} \nabla \tilde{f}_k(\theta_{k,e}^t; \xi_{k,e}) + \delta_{\text{total}}
\]

where $\|\delta_{\text{total}}\| \leq \varepsilon_{WB}$ is the Wasserstein Barycenter approximation error.

Taking Norm Squared, we get below:

\[
\begin{aligned}
\|\theta_{t+1} - \theta_t\|^2 &= \left\| -\eta \cdot \frac{1}{K} \sum_{k=1}^K \sum_{e=0}^{E-1} \nabla \tilde{f}_k(\theta_{k,e}^t; \xi_{k,e}) + \delta_{\text{total}} \right\|^2 \\
\end{aligned}
\]

Now we apply the below inequality:

\[
\|a + b\|^2 \leq 2\|a\|^2 + 2\|b\|^2
\]

So, we get below:

\[
\begin{aligned}
\|\theta_{t+1} - \theta_t\|^2 &\leq 2\left\| \eta \cdot \frac{1}{K} \sum_{k=1}^K \sum_{e=0}^{E-1} \nabla \tilde{f}_k(\theta_{k,e}^t; \xi_{k,e}) \right\|^2 + 2\|\delta_{\text{total}}\|^2 \\
\end{aligned}
\]

Now we take Expectation of the above inequality:

\[
\begin{aligned}
\label{jj}
\mathbb{E}\left[\|\theta_{t+1} - \theta_t\|^2\right] &\leq 2\mathbb{E}\left[\left\| \eta \cdot \frac{1}{K} \sum_{k=1}^K \sum_{e=0}^{E-1} \nabla \tilde{f}_k(\theta_{k,e}^t; \xi_{k,e}) \right\|^2\right] + 2\varepsilon_{WB}^2
\end{aligned}
\]


Using Jensen's inequality and bounded gradients, we obtain Eq.~\ref{qqqq}:
\begin{equation}
\begin{aligned}
&\leq 2\eta^2 \mathbb{E}\left[\left\|\frac{1}{K} \sum_{k=1}^K \sum_{e=0}^{E-1} \nabla \tilde{f}_k(\theta^{k,e}_t; \xi_{k,e})\right\|^2\right] + 2\varepsilon_{WB}^2 \\
&\leq 2\eta^2 E^2 G^2 + 2\varepsilon_{WB}^2
\label{qqqq}
\end{aligned}
\end{equation}

From Lemma 3, we can derive the following.
\begin{align}
\mathbb{E}[\|\theta^{k,e}_t - \theta_t\|^2] \leq \frac{\eta^2 e^2 \tilde{\sigma}^2}{2} + \frac{\eta^2 e^2 \tilde{L}^2 G^2}{2}.
\end{align}

Summing over all local steps yields:
\begin{equation}
\label{ncvb}
\begin{aligned}
\frac{1}{K} \sum_{k=1}^K \sum_{e=0}^{E-1} \mathbb{E}[\|\theta^{k,e}_t - \theta_t\|^2] &\leq \frac{\eta^2 \tilde{\sigma}^2}{2K} \sum_{e=0}^{E-1} e^2 + \frac{\eta^2 \tilde{L}^2 G^2}{2K} \sum_{e=0}^{E-1} e^2 \\
&\leq \frac{\eta^2 E^2 \tilde{\sigma}^2}{2K} + \frac{\eta^2 E^2 \tilde{L}^2 G^2}{2K}.
\end{aligned}
\end{equation}

We use $\sum_{e=0}^{E-1} e^2 \leq E^3/3 \leq E^2$ for practical purposes. Substituting Eq.~\ref{ncvb} into Eq.~\ref{fjx}, we get Eq.~\ref{ddf}:

\begin{equation}
\begin{aligned}
\label{ddf}
&\mathbb{E}[\tilde{F}(\theta_{t+1})] \leq \mathbb{E}[\tilde{F}(\theta_t)] - \frac{\eta}{2} \mathbb{E}[\|\nabla \tilde{F}(\theta_t)\|^2] 
 + \frac{\eta \tilde{L}^2 E}{4K} \left( \frac{\eta^2 E^2 \tilde{\sigma}^2}{2K} + \frac{\eta^2 E^2 \tilde{L}^2 G^2}{2K} \right) + \frac{\eta E \tilde{\sigma}^2}{4} \\
&\quad + \frac{\tilde{L}}{2} (2\eta^2 E^2 G^2 + 2\varepsilon_{WB}^2) + G\varepsilon_{WB}.
\end{aligned}
\end{equation}

Simplifying the higher-order terms and keeping dominant terms:
\begin{align}
\label{5bnm}
\mathbb{E}[\tilde{F}(\theta_{t+1})] \leq \mathbb{E}[\tilde{F}(\theta_t)] - \frac{\eta}{2} \mathbb{E}[\|\nabla \tilde{F}(\theta_t)\|^2] + \frac{\eta \tilde{L}E \tilde{\sigma}^2}{K} + \eta \tilde{L}E^2 G^2 + \varepsilon_{WB} G.
\end{align}

From the definition of convex functions, we have:
\begin{align}
\langle \nabla \tilde{F}(\theta_t), \theta_t - \theta^* \rangle \geq \tilde{F}(\theta_t) - \tilde{F}(\theta^*)
\end{align}

Applying the Cauchy–Schwarz inequality along with $2ab \leq a^2 + b^2$, we obtain:
\begin{align}
\|\nabla \tilde{F}(\theta_t)\|^2 \geq \frac{2(\tilde{F}(\theta_t) - \tilde{F}(\theta^*))\langle \nabla \tilde{F}(\theta_t), \theta_t - \theta^* \rangle}{\|\theta_t - \theta^*\|^2}
\end{align}

Subtracting $\tilde{F}(\theta^*)$ from both sides of Eq.~\ref{5bnm}, we obtain:
\begin{align}
\mathbb{E}[\tilde{F}(\theta_{t+1})] - \tilde{F}(\theta^*) \leq \mathbb{E}[\tilde{F}(\theta_t)] - \tilde{F}(\theta^*) - \frac{\eta}{2} \mathbb{E}[\|\nabla \tilde{F}(\theta_t)\|^2] + C,
\end{align}

where $C = \frac{\eta \tilde{L}E \tilde{\sigma}^2}{K} + \eta \tilde{L}E^2 G^2 + \varepsilon_{WB} G$.

For convex functions, we leverage the inequality:
\begin{align}
\mathbb{E}[\|\nabla \tilde{F}(\theta_t)\|^2] \geq \frac{2(\mathbb{E}[\tilde{F}(\theta_t)] - \tilde{F}(\theta^*))}{\eta}.
\end{align}

This yields the recursive bound:
\begin{align}
\mathbb{E}[\tilde{F}(\theta_{t+1})] - \tilde{F}(\theta^*) \leq \frac{1}{2}(\mathbb{E}[\tilde{F}(\theta_t)] - \tilde{F}(\theta^*)) + C.
\end{align}

Unrolling the recurrence gives:
\begin{align}
\mathbb{E}[\tilde{F}(\theta_T)] - \tilde{F}(\theta^*) \leq \left(\frac{1}{2}\right)^T (\tilde{F}(\theta_0) - \tilde{F}(\theta^*)) + 2C.
\end{align}

Using Jensen's inequality for the averaged iterate $\bar{\theta}_T = \frac{1}{T}\sum_{t=0}^{T-1} \theta_t$:
\begin{equation}
\begin{aligned}
\mathbb{E}[\tilde{F}(\bar{\theta}_T)] - \tilde{F}(\theta^*) &\leq \frac{1}{T}\sum_{t=0}^{T-1} (\mathbb{E}[\tilde{F}(\theta_t)] - \tilde{F}(\theta^*)) \\
&\leq \frac{2\|\theta_0 - \theta^*\|^2}{\eta T} + \frac{\eta \tilde{L}E \tilde{\sigma}^2}{K} + \eta \tilde{L}E^2 G^2 + \varepsilon_{WB} G
\end{aligned}
\end{equation}

Putting $\eta = \frac{1}{4\tilde{L}E}$ in above equation and simplifying yields:\\

\begin{align}
\mathbb{E}[\tilde{F}(\bar{\theta}_T) - \tilde{F}(\theta^*)] \leq \frac{8 \tilde{L} E \|\theta_0 - \theta^*\|^2}{T} + \frac{\tilde{\sigma}^2}{4K} + \frac{E G^2}{4} + \epsilon_{WB} G
\end{align}

\begin{align}
\mathbb{E}[\tilde{F}(\bar{\theta}_T) - \tilde{F}(\theta^*)]
= \mathcal{O} \left( \frac{1}{{T}} \right)  + constant
\end{align}

\subsection{Non-Convex Convergence Analysis}

\label{thm:nonconvex_convergence}
\textbf{Theorem 2:}
Given that Assumptions 1–6 are satisfied and $\tilde{F}(\theta)$ is non-convex, setting the learning rate $\eta \leq \frac{1}{4\tilde{L}E}$ ensures that FedDUAL achieves the following convergence guarantee:

\begin{align}
\frac{1}{T} \sum_{t=0}^{T-1} \mathbb{E}[\|\nabla \tilde{F}(\theta_t)\|^2] &\leq \frac{2(\tilde{F}(\theta_0) - \tilde{F}(\theta^*))}{\eta T} + \frac{\eta \tilde{L}E \tilde{\sigma}^2}{K}+ \eta \tilde{L}E^2 G^2 + \frac{\varepsilon_{WB} G}{\eta}, \label{eq:nonconvex_bound}
\end{align}
where $\theta^*$ is any global minimum of $\tilde{F}(\theta)$ and $\tilde{\sigma}^2$ is as defined in Lemma 2.

\textbf{Proof:}
From the $\tilde{L}$-smoothness of $\tilde{F}(\theta)$ (established in Lemma 1), we have:
\begin{align}
\tilde{F}(\theta_{t+1}) &\leq \tilde{F}(\theta_t) + \langle\nabla \tilde{F}(\theta_t), \theta_{t+1} - \theta_t\rangle + \frac{\tilde{L}}{2}\|\theta_{t+1} - \theta_t\|^2. 
\label{eq:smoothness_nonconvex}
\end{align}

The FedDUAL update rule is defined as follows, as presented in Eq.~\ref{u89}:


\begin{align}
\theta_{t+1} - \theta_t = -\frac{\eta}{K} \sum_{k=1}^K \sum_{e=0}^{E-1} \nabla \tilde{f}_k(\theta_t^{k,e}; \xi^{k,e}) + \delta_{total}, \label{eq:feddual_update_nonconvex}
\end{align}
where $\|\delta_{total}\| \leq \varepsilon_{WB}$ represents the Wasserstein Barycenter approximation error.

Substituting Eq.~\ref{eq:feddual_update_nonconvex} into the inner product in Eq.~\ref{eq:smoothness_nonconvex}, we get Eq.~\ref{eq:inner_product_decomp}.
\begin{align}
&\langle\nabla \tilde{F}(\theta_t), \theta_{t+1} - \theta_t\rangle 
= -\frac{\eta}{K} \sum_{k=1}^K \sum_{e=0}^{E-1} \langle\nabla \tilde{F}(\theta_t), \nabla \tilde{f}_k(\theta_t^{k,e}; \xi^{k,e})\rangle + \langle\nabla \tilde{F}(\theta_t), \delta_{total}\rangle \label{eq:inner_product_decomp}
\end{align}

Applying the Cauchy–Schwarz inequality to the error term in Eq.~\ref{eq:inner_product_decomp}, we obtain the following.
\begin{align}
|\langle\nabla \tilde{F}(\theta_t), \delta_{total}\rangle| \leq \|\nabla \tilde{F}(\theta_t)\|\|\delta_{total}\| \leq G\varepsilon_{WB}. \label{eq:error_term_bound}
\end{align}


For each gradient term in Eq.~\ref{eq:inner_product_decomp}, we add and subtract the term $\nabla \tilde{f}_k(\theta_t)$:
\begin{align}
&\langle\nabla \tilde{F}(\theta_t), \nabla \tilde{f}_k(\theta_t^{k,e}; \xi^{k,e})\rangle
= \langle\nabla \tilde{F}(\theta_t), \nabla \tilde{f}_k(\theta_t)\rangle + \langle\nabla \tilde{F}(\theta_t), \nabla \tilde{f}_k(\theta_t^{k,e}; \xi^{k,e}) - \nabla \tilde{f}_k(\theta_t)\rangle 
\label{eq:gradient_decomp}
\end{align}

The first term of Eq.~\ref{eq:gradient_decomp} gives the following:
\begin{align}
\frac{1}{K} \sum_{k=1}^K \langle\nabla \tilde{F}(\theta_t), \nabla \tilde{f}_k(\theta_t)\rangle = \|\nabla \tilde{F}(\theta_t)\|^2. \label{eq:gradient_squared_norm}
\end{align}

For the second term in Eq.~\ref{eq:gradient_decomp}, we further decompose:
\begin{align}
&\nabla \tilde{f}_k(\theta_t^{k,e}; \xi^{k,e}) - \nabla \tilde{f}_k(\theta_t)= [\nabla \tilde{f}_k(\theta_t^{k,e}; \xi^{k,e}) - \nabla \tilde{f}_k(\theta_t^{k,e})] + [\nabla \tilde{f}_k(\theta_t^{k,e}) - \nabla \tilde{f}_k(\theta_t)]. \label{eq:deviation_decomp}
\end{align}


Taking expectations over the second term in Eq.~\ref{eq:gradient_decomp} and applying Young's inequality with parameter $\alpha$, we derive (omitting details identical to the convex setting):
\begin{align}
\mathbb{E}\left[|\langle\nabla \tilde{F}(\theta_t), \nabla \tilde{f}_k(\theta_t^{k,e}; \xi^{k,e}) - \nabla \tilde{f}_k(\theta_t)\rangle|\right] 
&\leq \frac{\mathbb{E}[\|\nabla \tilde{F}(\theta_t)\|^2]}{2\alpha} + \frac{\alpha}{2}\mathbb{E}[\|\nabla \tilde{f}_k(\theta_t^{k,e}; \xi^{k,e}) - \nabla \tilde{f}_k(\theta_t^{k,e})\|^2] \nonumber \\
&\quad + \frac{\alpha \tilde{L}^2}{2}\mathbb{E}[\|\theta_t^{k,e} - \theta_t\|^2]. 
\label{eq:young_bound}
\end{align}

Using Lemma 2 for the variance bound and setting $\alpha = \frac{1}{2}$ in Eq.~\ref{eq:young_bound}, we obtain the following deviation bound:
\begin{align}
&\mathbb{E}[|\langle\nabla \tilde{F}(\theta_t), \nabla \tilde{f}_k(\theta_t^{k,e}; \xi^{k,e}) - \nabla \tilde{f}_k(\theta_t)\rangle|] 
\leq \mathbb{E}[\|\nabla \tilde{F}(\theta_t)\|^2] + \frac{\tilde{\sigma}^2}{4} + \frac{\tilde{L}^2}{4}\mathbb{E}[\|\theta_t^{k,e} - \theta_t\|^2]. 
\label{eq:deviation_bound}
\end{align}

By combining all the terms and substituting them into Eq.~\ref{eq:inner_product_decomp}, we get the following bound:
\begin{align}
&\mathbb{E}[\langle\nabla \tilde{F}(\theta_t), \theta_{t+1} - \theta_t\rangle] 
\leq -\eta\mathbb{E}[\|\nabla \tilde{F}(\theta_t)\|^2] + \frac{\eta}{K}\sum_{k=1}^K \sum_{e=0}^{E-1} \left[\mathbb{E}[\|\nabla \tilde{F}(\theta_t)\|^2] + \frac{\tilde{\sigma}^2}{4} + \frac{\tilde{L}^2}{4}\mathbb{E}[\|\theta_t^{k,e} - \theta_t\|^2]\right] \nonumber \\
&\quad + G\varepsilon_{WB}. 
\label{eq:combined_bound}
\end{align}


Simplifying Eq.~\ref{eq:combined_bound}, we arrive at:
\begin{align}
&\mathbb{E}[\langle\nabla \tilde{F}(\theta_t), \theta_{t+1} - \theta_t\rangle]
&\leq -\frac{\eta}{2}\mathbb{E}[\|\nabla \tilde{F}(\theta_t)\|^2] + \frac{\eta \tilde{L}^2 E}{4K}\sum_{k=1}^K \sum_{e=0}^{E-1} \mathbb{E}[\|\theta_t^{k,e} - \theta_t\|^2] + \frac{\eta E \tilde{\sigma}^2}{4} + G\varepsilon_{WB}.
\label{eq:simplified_bound}
\end{align}


From Lemma 3 and using the fact that $\sum_{e=0}^{E-1} e^2 \leq E^3/3 \leq E^2$ for practical purposes, we obtain:
\begin{align}
\frac{1}{K}\sum_{k=1}^K \sum_{e=0}^{E-1} \mathbb{E}[\|\theta_t^{k,e} - \theta_t\|^2] &\leq \frac{\eta^2 E^2 \tilde{\sigma}^2}{2K} + \frac{\eta^2 E^2 \tilde{L}^2 G^2}{2K} \label{eq:local_drift_bound}
\end{align}


From Eq.~\ref{eq:feddual_update_nonconvex}, applying Jensen's inequality yields:
\begin{align}
\mathbb{E}[\|\theta_{t+1} - \theta_t\|^2] &\leq 2\mathbb{E}\left[\left\|\frac{\eta}{K}\sum_{k=1}^K \sum_{e=0}^{E-1} \nabla \tilde{f}_k(\theta_t^{k,e}; \xi^{k,e})\right\|^2\right] + 2\varepsilon_{WB}^2 
\leq 2\eta^2 E^2 G^2 + 2\varepsilon_{WB}^2.
\label{eq:param_update_bound}
\end{align}


By combining all the above terms and substituting them into the smoothness inequality in Eq.~\ref{eq:smoothness_nonconvex}, we obtain:

\begin{align}
\mathbb{E}[\tilde{F}(\theta_{t+1})] &\leq \mathbb{E}[\tilde{F}(\theta_t)] - \frac{\eta}{2}\mathbb{E}[\|\nabla \tilde{F}(\theta_t)\|^2] + \frac{\eta \tilde{L}E \tilde{\sigma}^2}{K} + \eta \tilde{L}E^2 G^2 + \frac{\tilde{L}}{2}(2\eta^2 E^2 G^2 + 2\varepsilon_{WB}^2) + G\varepsilon_{WB}.
\label{eq:final_smoothness}
\end{align}

Rearranging and using the learning rate condition $\eta \leq \frac{1}{4\tilde{L}E}$ in Eq.~\ref{eq:final_smoothness}, we get:
\begin{align}
\frac{\eta}{2}\mathbb{E}[\|\nabla \tilde{F}(\theta_t)\|^2] &\leq \mathbb{E}[\tilde{F}(\theta_t)] - \mathbb{E}[\tilde{F}(\theta_{t+1})] + \frac{\eta \tilde{L}E \tilde{\sigma}^2}{K} + \eta \tilde{L}E^2 G^2 
 + \tilde{L}\varepsilon_{WB}^2 + G\varepsilon_{WB}.
 \label{eq:rearranged}
\end{align}

Summing the above equation over $t = 0, 1, \ldots, T-1$ and dividing by $\frac{\eta T}{2}$:
\begin{align}
\frac{1}{T} \sum_{t=0}^{T-1} \mathbb{E}[\|\nabla \tilde{F}(\theta_t)\|^2] &\leq \frac{2(\tilde{F}(\theta_0) - \mathbb{E}[\tilde{F}(\theta_T)])}{\eta T} + \frac{\eta \tilde{L}E \tilde{\sigma}^2}{K} + \eta \tilde{L}E^2 G^2 
 + \frac{2\tilde{L}\varepsilon_{WB}^2}{\eta} + \frac{2G\varepsilon_{WB}}{\eta}.
 \label{eq:telescoping}
\end{align}

Since $\tilde{F}(\theta_T) \geq \tilde{F}(\theta^*)$, and combining the error terms, we have:
\begin{align}
\frac{1}{T} \sum_{t=0}^{T-1} \mathbb{E}[\|\nabla \tilde{F}(\theta_t)\|^2] &\leq \frac{2(\tilde{F}(\theta_0) - \tilde{F}(\theta^*))}{\eta T} + \frac{\eta \tilde{L}E \tilde{\sigma}^2}{K} + \eta \tilde{L}E^2 G^2 + \frac{\varepsilon_{WB} G}{\eta},
\label{conv}
\end{align}

where the term $\tilde{L}\varepsilon_{WB}^2$ has been absorbed into the dominant $\frac{\varepsilon_{WB} G}{\eta}$ term for simplicity. To minimize the right-hand side, we choose the learning rate $\eta$ that balances the first and second terms:

\[
\eta = \sqrt{\frac{2 \left( \tilde{F}(\theta_0) - \tilde{F}(\theta^*) \right)}{\tilde{L} \tilde{\sigma}^2 T}}.
\]


Substituting this choice of $\eta$ into the inequality gives:
\[
\frac{1}{T} \sum_{t=0}^{T-1} \mathbb{E} \left[ \|\nabla \tilde{F}(\theta_t)\|^2 \right] 
\leq 
\frac{2 \left(\tilde{F}(\theta_0) - \tilde{F}(\theta^*) \right)}{\sqrt{ \frac{2 \left( \tilde{F}(\theta_0) - \tilde{F}(\theta^*) \right)}{\tilde{L} \tilde{\sigma}^2 T} } \cdot T}
+ \tilde{L} \tilde{\sigma}^2 \cdot \sqrt{ \frac{2 \left( \tilde{F}(\theta_0) - \tilde{F}(\theta^*) \right)}{\tilde{L} \tilde{\sigma}^2 T} }
+ \epsilon_{\text{WB}} G.
\]

Simplifying both terms, we obtain:
\[
\frac{1}{T} \sum_{t=0}^{T-1} \mathbb{E} \left[ \|\nabla \tilde{F}(\theta_t)\|^2 \right] 
\leq 
\frac{2 \sqrt{2 (\tilde{F}(\theta_0) - \tilde{F}(\theta^*)) \tilde{L} \tilde{\sigma}^2}}{\sqrt{T}} 
+ \frac{ \sqrt{2 (\tilde{F}(\theta_0) - \tilde{F}(\theta^*)) \tilde{L} \tilde{\sigma}^2} }{ \sqrt{T} } 
+ \epsilon_{\text{WB}} G
\]


\[
\frac{1}{T} \sum_{t=0}^{T-1} \mathbb{E} \left[ \|\nabla \tilde{F}(\theta_t)\|^2 \right] 
\leq 
\frac{3 \sqrt{2 (\tilde{F}(\theta_0) - \tilde{F}(\theta^*)) \tilde{L} \tilde{\sigma}^2}}{\sqrt{T}} 
+ \epsilon_{\text{WB}} G.
\]


Therefore, the convergence rate is:

\[
\frac{1}{T} \sum_{t=0}^{T-1} \mathbb{E} \left[ \|\nabla \tilde{F}(\theta_t)\|^2 \right] 
= \mathcal{O} \left( \frac{1}{\sqrt{T}} \right) + \epsilon_{\text{WB}} G
\]


\subsection{Discussion on Convergence Bounds}

Our theoretical analysis establishes convergence guarantees for both convex and non-convex settings:

\begin{itemize}
    \item \textbf{Convex Setting:} Theorem~1 shows that FedDUAL achieves
    \[
    \mathbb{E}[F(\bar{\theta}_T) - F(\theta^*)] \leq \mathcal{O}\left(\frac{1}{T}\right) + \text{constant},
    \]
    which matches the optimal rate for first-order methods under smoothness and convexity assumptions. This indicates that FedDUAL preserves convergence efficiency while incorporating dynamic weighting and Wasserstein aggregation.
    
    \item \textbf{Non-Convex Setting:} Theorem~2 proves that
    \[
    \frac{1}{T}\sum_{t=0}^{T-1}\mathbb{E}\big[\|\nabla F(\theta_t)\|^2\big] \leq \mathcal{O}\left(\frac{1}{\sqrt{T}}\right) + \epsilon_{\text{WB}}G,
    \]
    consistent with the standard rate for non-convex optimization in federated learning. The additional $\epsilon_{\text{WB}}$ term quantifies the effect of Wasserstein Barycenter approximation, ensuring that its impact remains bounded.
\end{itemize}

\paragraph{Comparison with state of the art methods:} Standard algorithms such as FedAvg~\citep{mcmahan2017communication} and FedProx~\citep{li2020federated} achieve similar asymptotic rates ($\mathcal{O}(1/T)$ for convex and $\mathcal{O}(1/\sqrt{T})$ for non-convex), but suffer from significant degradation due to client drift under non-IID data. Specifically, FedAvg incurs heterogeneity-dependent terms of $\mathcal{O}(\zeta^2 K^2/T)$ in non-convex settings, while FedProx improves this to $\mathcal{O}(\zeta^2 K/T)$ through proximal regularization. Variance-reduced methods like SCAFFOLD~\citep{karimireddy2020scaffold} achieve superior bounds with heterogeneity terms of only $\mathcal{O}(\zeta^2/T)$ (independent of local steps $K$) and variance scaling as $\mathcal{O}(1/nKT)$, but require maintaining control variates at each client. Our bounds show that FedDUAL matches these theoretical rates while introducing Wasserstein aggregation and adaptive loss weighting, providing robustness to heterogeneity without the additional memory and communication overhead of control variates where $T$ denotes the number of global communication rounds, $K$ is the number of local SGD steps performed at each client between communications, $n$ is the total number of participating clients, $\zeta$ is the client drift parameter measuring data heterogeneity, n is the total clients, and $\eta$ is the learning rate.

\section{Computational Complexity Analysis}
\label{xdbb}
We present a formal complexity analysis comparing the proposed method with baseline approaches for a single communication round involving $K$ clients, a model of dimension $d$, and $n$ denoting the number of parameters in the final layer.

\subsubsection*{Client-Side Complexity}

\textbf{FedAvg:} The computational complexity per communication round is $\mathcal{O}(E \cdot B \cdot d)$, where $E$ denotes the number of local epochs, $B$ the batch size, and $d$ the model dimension.
This cost primarily arises from the forward and backward passes performed during local training on each client.\\
\textbf{FedProx:} The per-round computational complexity is $\mathcal{O}(E \cdot B \cdot d + d)$, where the additional $\mathcal{O}(d)$ term accounts for computing the proximal regularization term $|\theta_k - \theta_g|^2$.\\
\textbf{SCAFFOLD:} The per-round complexity is $\mathcal{O}(E \cdot B \cdot d + 2d)$, with the extra $\mathcal{O}(d)$ arising from storing and updating control variates used to correct client drift.\\
\textbf{FedDUAL (Proposed):} The per-round complexity is $\mathcal{O}(E \cdot B \cdot d + d)$, where the $\mathcal{O}(d)$ term corresponds to KL divergence computation comprising weight flattening, softmax operations, and divergence calculation, all linear in $d$. Thus, FedDUAL maintains the same asymptotic complexity as FedProx.\\
Please note that Our adaptive loss adds negligible client-side overhead compared to local training cost (as $E \cdot B \cdot d \gg d$).

\subsubsection*{Server-Side Complexity}

\textbf{FedAvg:} The aggregation complexity per communication round is $\mathcal{O}(K \cdot d)$, corresponding to the simple averaging of model parameters across $K$ clients.
\textbf{FedProx/FedNova:} The aggregation step also has complexity $\mathcal{O}(K \cdot d)$, as weight normalization and scaling operations remain linear in the model dimension.\\
\textbf{SCAFFOLD:} The aggregation complexity is $\mathcal{O}(K \cdot d)$, dominated by the averaging of client updates along with control variate terms.\\
\textbf{FedDUAL (Proposed):} The aggregation complexity is $\mathcal{O}(K \cdot d + I \cdot K \cdot n^2)$, where $\mathcal{O}(K \cdot d)$ accounts for standard aggregation of lower-layer parameters and $\mathcal{O}(I \cdot K \cdot n^2)$ arises from computing the Wasserstein barycenter for the final layers. Here, $I$ denotes the number of Sinkhorn iterations (typically 100–150), and $n$ represents the number of parameters in the last layers ($n \ll d$); for instance, in VGG-16, $d \approx 138$M total parameters, where the last two fully connected layers contain $n \approx 67.2$M parameters (two layers of 4096$\times$4096 each), constituting approximately 48.7\% of the total parameters.

Observation: While $n$ represents a substantial portion of the network, the Wasserstein aggregation complexity $\mathcal{O}(I \cdot K \cdot n^2)$ remains tractable when applied selectively. For our setup:
\begin{itemize}
 \item VGG-16 on CIFAR-10: $n = 67{,}240{,}000$ (last two FC layers), $I = 150$, $K = 10$ (per round)
 \item Wasserstein aggregation: $\sim 150 \times 10 \times (67{,}240{,}000)^2 \approx 6.78 \times 10^{17}$ operations
\item Standard aggregation: $10 \times 138 \times 10^6 \approx 1.38 \times 10^{9}$ operations
 \item Overhead ratio: $\sim 4.9 \times 10^{8}\times$ for the affected layers
\end{itemize}
However, this computational overhead is incurred only during the aggregation phase on the server, not during client-side training. Moreover, practical implementations use approximate Wasserstein distance computations (e.g., Sinkhorn iterations) which significantly reduce this theoretical complexity while maintaining the benefits of permutation-invariant aggregation.

subsubsection*{Amortized Analysis}
Considering total time-to-convergence:
\begin{equation}
T_{\text{total}} = R \times (T_{\text{client}} + T_{\text{server}})
\end{equation}
where $R$ is the number of rounds, $T_{\text{client}}$ is the per-round client training time, and $T_{\text{server}}$ is the per-round server aggregation time.

Since FedDUAL achieves convergence in $\sim$40\% fewer rounds (refer to Fig.~\ref{fig4}) while incurring $\sim$35\% server-side overhead per round:
\begin{equation}
\frac{T_{\text{FedDUAL}}}{T_{\text{FedAvg}}} = \frac{0.6R \times (T_{\text{client}} + 1.35 \times T_{\text{server}})}{R \times (T_{\text{client}} + T_{\text{server}})} = 0.6 \times \left(1 + \frac{0.35 \times T_{\text{server}}}{T_{\text{client}} + T_{\text{server}}}\right).
\end{equation}

In typical FL scenarios where client training dominates ($T_{\text{client}} \gg T_{\text{server}}$), we have $\frac{T_{\text{server}}}{T_{\text{client}} + T_{\text{server}}} \approx 0$, yielding:
\begin{equation}
\frac{T_{\text{FedDUAL}}}{T_{\text{FedAvg}}} \approx 0.6 \times (1 + 0) = 0.6.
\end{equation}
Thus, the 35\% server overhead becomes negligible, and FedDUAL achieves approximately 40\% reduction in total wall-clock time due to faster convergence.

To empirically validate the assumption that $T_{client} \gg T_{server}$, we measured the average time required for a single client update and the total server update time per round. The results, summarized in Table~\ref{runtime2}, corroborate the assumption, demonstrating that client-side updates dominate the overall computational cost which further validate the general assumption made above.

\begin{table}[ht]
\caption{Wall clock time (in seconds) of the proposed method and the Fedavg algorithm for client and server side update. }
\centering
\scalebox{1}{
\begin{tabular}{lcccc}
\toprule
& \multicolumn{2}{c}{CIFAR-10} & \multicolumn{2}{c}{FMNIST} \\
\cmidrule(lr){2-3} \cmidrule(lr){4-5}
& T\_{client} & T\_{server} & T\_{client} & T\_{server} \\
\midrule
FedAvg & 17.71 & 0.44 & 2.14 & 0.05 \\
Proposed & 27.49 &  1.94 & 2.62 & 1.95 \\
\bottomrule
\end{tabular}}
\label{runtime2}
\end{table}

\section{Runtime Analysis}
\label{run}
Table~\ref{runtime} presents the wall-clock training time (in hours) of the proposed method compared to several widely used federated learning baselines on the CIFAR-10 and FMNIST datasets. This time is recorded for fixed number of rounds for all the algorithms (80 rounds for CIFAR10 dataset and 180 rounds for FMNIST dataset). As expected, the runtime varies across algorithms due to differences in their communication strategies, local computation overhead, and auxiliary regularization terms. Among the baselines, FedAvg achieves the lowest runtime, since it performs simple model averaging without any additional constraints or control variates. FedProx and FedBN incur a slight increase in runtime due to the introduction of proximal terms and batch normalization handling at the client side, respectively. FedNova and SCAFFOLD exhibit moderately higher runtimes, attributed to gradient normalization and control variate updates. MOON further increases computational cost because of its contrastive representation alignment loss, while shows additional overhead due to maintaining both personalized and shared components during optimization. FedDyn records the highest runtime among the existing methods, as its dynamic regularization term requires per-round model adjustment and additional global parameter updates. The proposed method, though computationally more expensive than the baselines, remains efficient considering the performance benefits (Higher accuracy and faster convergence) it achieves. Specifically, it requires 5.33 hours on CIFAR-10 and 3.01 holurs on FMNIST, which is competitive with other advanced methods such as FedDyn and FedPVR, while providing superior convergence and generalization performance. Overall, the proposed method achieves a favorable balance between computational cost and performance gain, validating its practicality in real-world federated learning scenarios.

We additionally report the wall-clock time (in hours) required to reach target accuracies of 0.40 for CIFAR-10 and 0.70 for FMNIST across all baselines and the proposed method, as summarized in Table~\ref{tab:time_to_target}. The results demonstrate that the proposed method consistently achieves the target accuracy in less time than the baselines, thereby confirming that the faster convergence observed in rounds (see Fig.~\ref{fig4}) effectively translates into faster overall wall-clock convergence.


\begin{table}[ht]
\caption{Wall clock performance (in hours) of the proposed method and the baselines. * shows the respective method failed to converge.}
\centering
\scalebox{1}{
\begin{tabular}{lcc}
\toprule
& CIFAR-10 & FMNIST \\
\midrule
FedAvg & 2.66 & 1.75 \\
FedProx & 2.90 & 1.90 \\
FedNova & 3.10 & * \\
FedBN & * & 1.80 \\
FedDyn & 4.89 & 2.91 \\
MOON & 3.80 & 2.40 \\
SCAFFOLD & * & * \\
FedPVR & 4.50 & 2.90 \\
\midrule
\textbf{Proposed} & 5.33 & 3.01 \\
\bottomrule
\end{tabular}}
\label{runtime}
\end{table}

\begin{table}[ht]
\caption{Wall-clock time to reach target accuracy (in hours). * indicates that the respective method failed to converge.}
\centering
\scalebox{1}{
\begin{tabular}{lcc}
\toprule
& CIFAR-10 & FMNIST  \\
\midrule
FedAvg & 0.76 & 1.46 \\
FedProx & 1.61 & 1.90 \\
FedNova & 1.24 & * \\
FedBN & * & 1.80 \\
FedDyn & 4.89 & 2.91 \\
MOON & 1.27 & 1.47 \\
SCAFFOLD & * & * \\
FedPVR & 1.31 & 1.21 \\
\midrule
\textbf{Proposed} & \textbf{0.67} & \textbf{1.02} \\
\bottomrule
\end{tabular}}
\label{tab:time_to_target}
\end{table}

\section{Limitation and Future Work}
While the proposed FedDUAL framework achieves superior performance and faster convergence under severe data heterogeneity, where several state-of-the-art methods such as FedNova, FedBN, and SCAFFOLD fail to converge, it introduces higher computational cost at the server due to the iterative calculation of the Wasserstein Barycenter. On the client side, FedDUAL remains lightweight and is also more communication-efficient than methods like SCAFFOLD and FedPVR, as it only requires transmitting model updates without additional control variates. Although this extra computational overhead is limited to the server, which typically has sufficient resources, future research will focus on reducing the computational burden of the Wasserstein Barycenter calculation by developing more efficient algorithms, aiming to maintain or even improve the performance of the proposed framework. Moreover, FedDUAL is inherently compatible with standard privacy-preserving techniques. In particular, its adaptive loss and dynamic aggregation can be seamlessly integrated with Differential Privacy by adding noise to client updates prior to aggregation, without modifying the core design. Investigating this integration with formal privacy guarantees remains an important direction for future research.




\begin{figure*}
\centering
\includegraphics[width=0.8\textwidth]{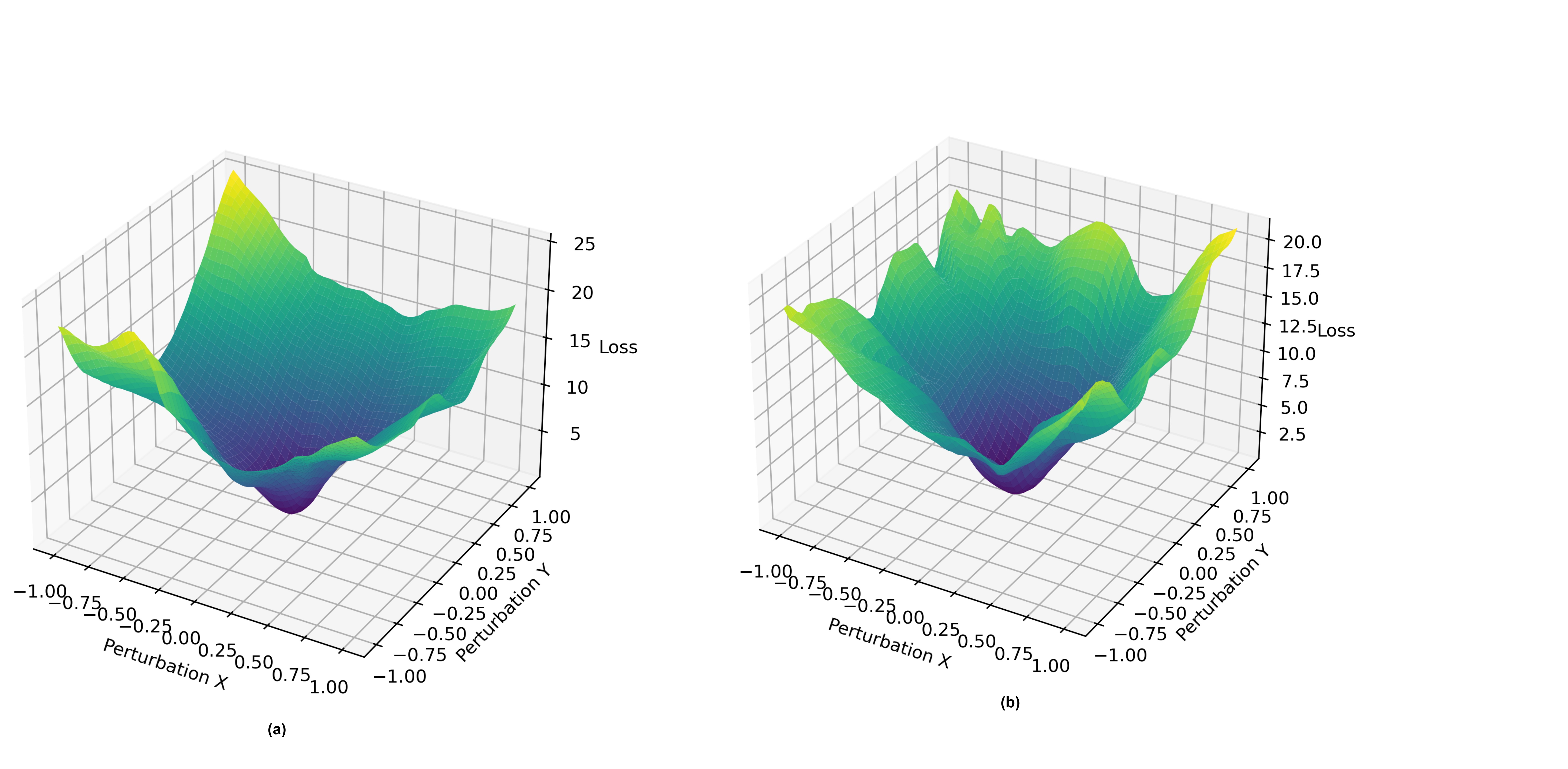}
\caption{Visualization of the loss surface for the global model trained on the FMNIST dataset with non-IID data ($\alpha=0.01$): (a) shows the loss surface for the global model trained using FedAvg, while (b) depicts the loss surface for the global model trained with the proposed method FedDUAL.}
\label{fig6}
\end{figure*}


\begin{figure}
\centering
\includegraphics[width=0.8\linewidth]{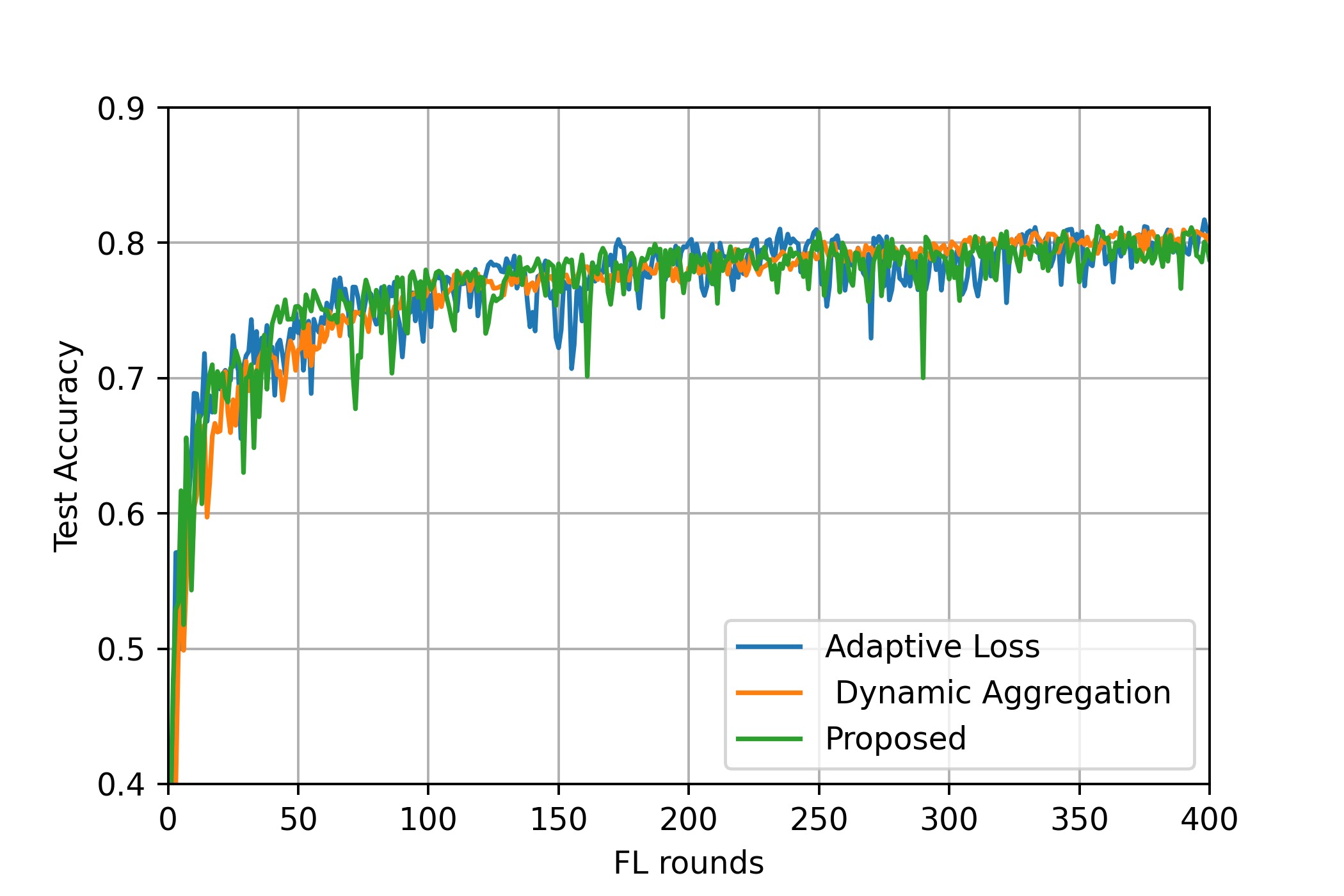}

\caption{Learning curves of the individual modules and the proposed method.}
\label{fig11}
\end{figure}

\begin{figure}
\centering
\includegraphics[width=0.7\linewidth]{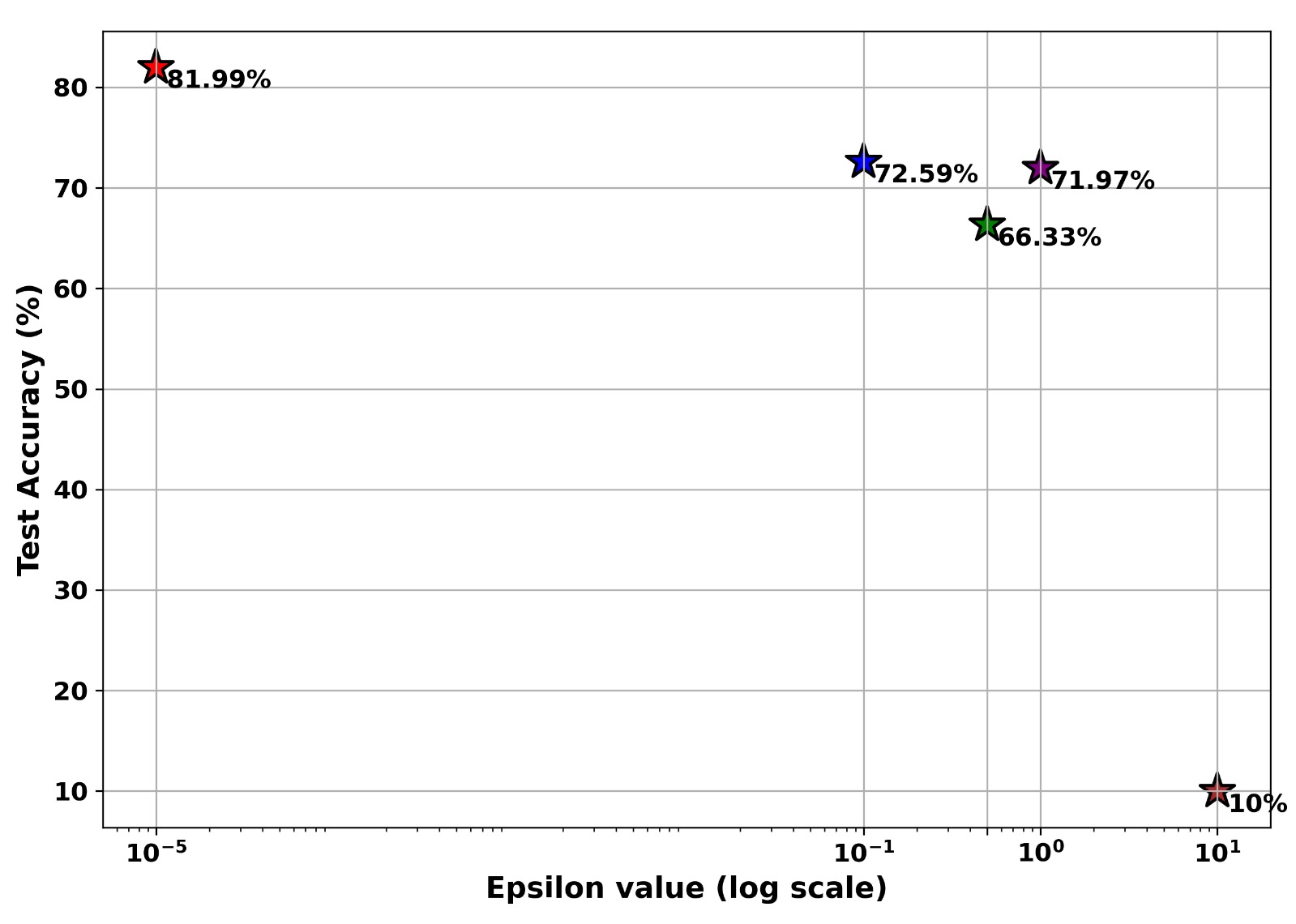}
\caption{Performance of the proposed method with different epsilon values.}
\label{fig9}
\end{figure}

\begin{figure}
\centering
\includegraphics[width=0.8\linewidth]{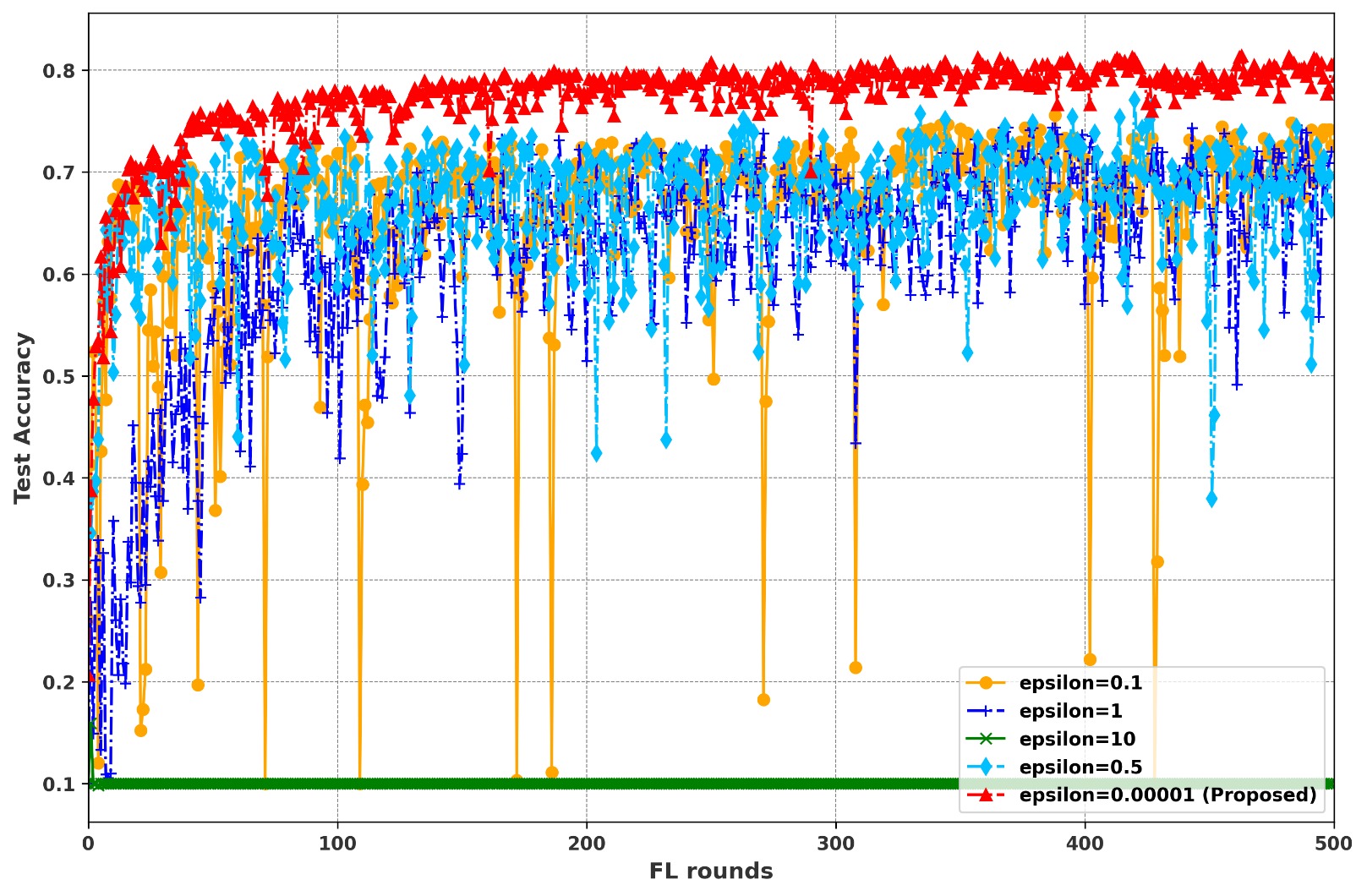}
\caption{Learning curve of the proposed method with different epsilon values.}
\label{fig10}
\end{figure}

\begin{figure}[t]
    \centering
    \includegraphics[width=\linewidth, height=5cm]{feddual_tmlr-error.jpg}
    \caption{Learning curves of the proposed method and baselines with error bars.}
    \label{13jhjh01}
\end{figure}


\begin{figure}
\centering
\includegraphics[width=0.8\linewidth]{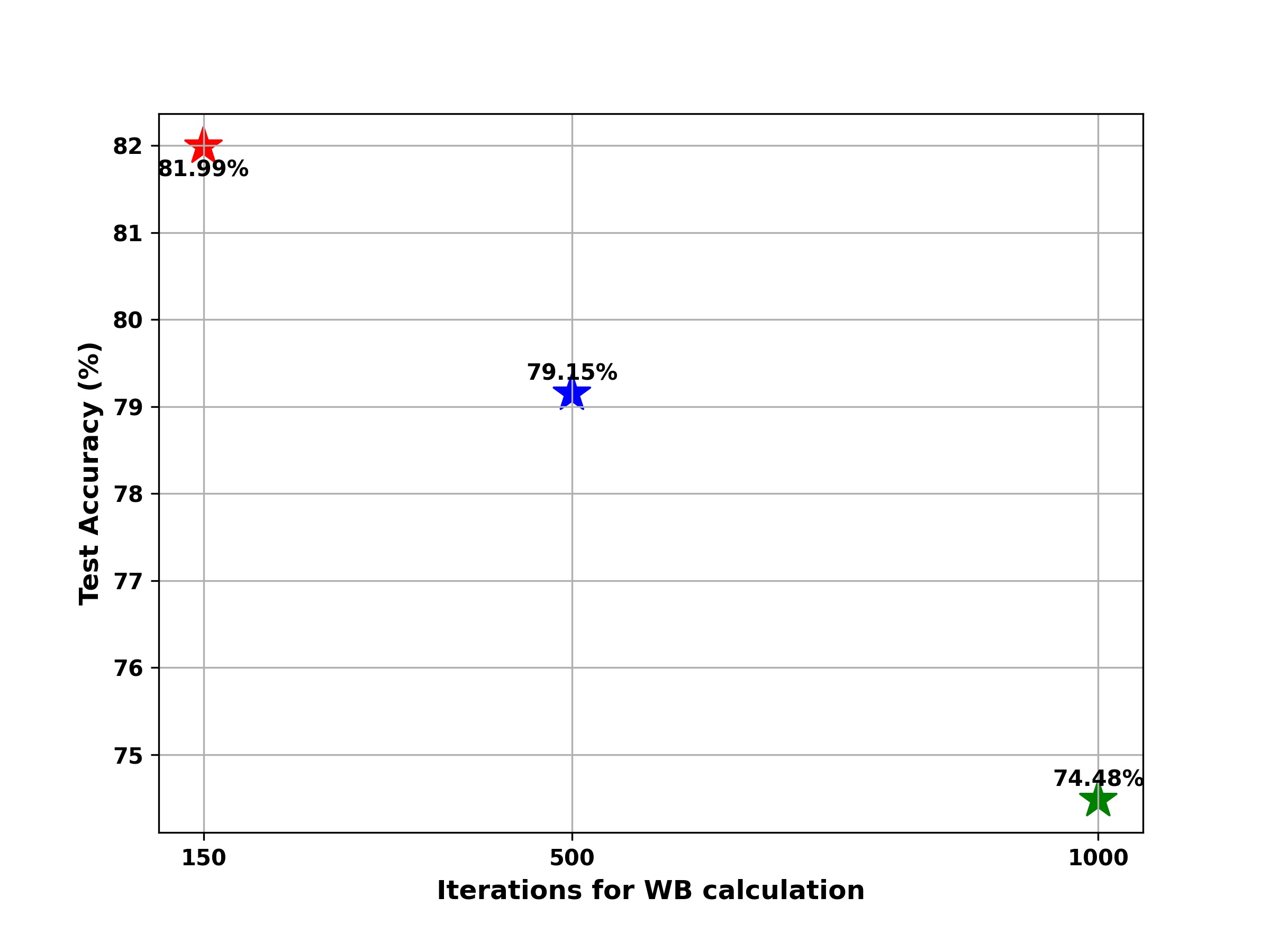}
\caption{Performance across different number of Iterations for WB calculation.}
\label{fig12}
\end{figure}

\begin{figure}
\centering
\includegraphics[width=0.8\linewidth]{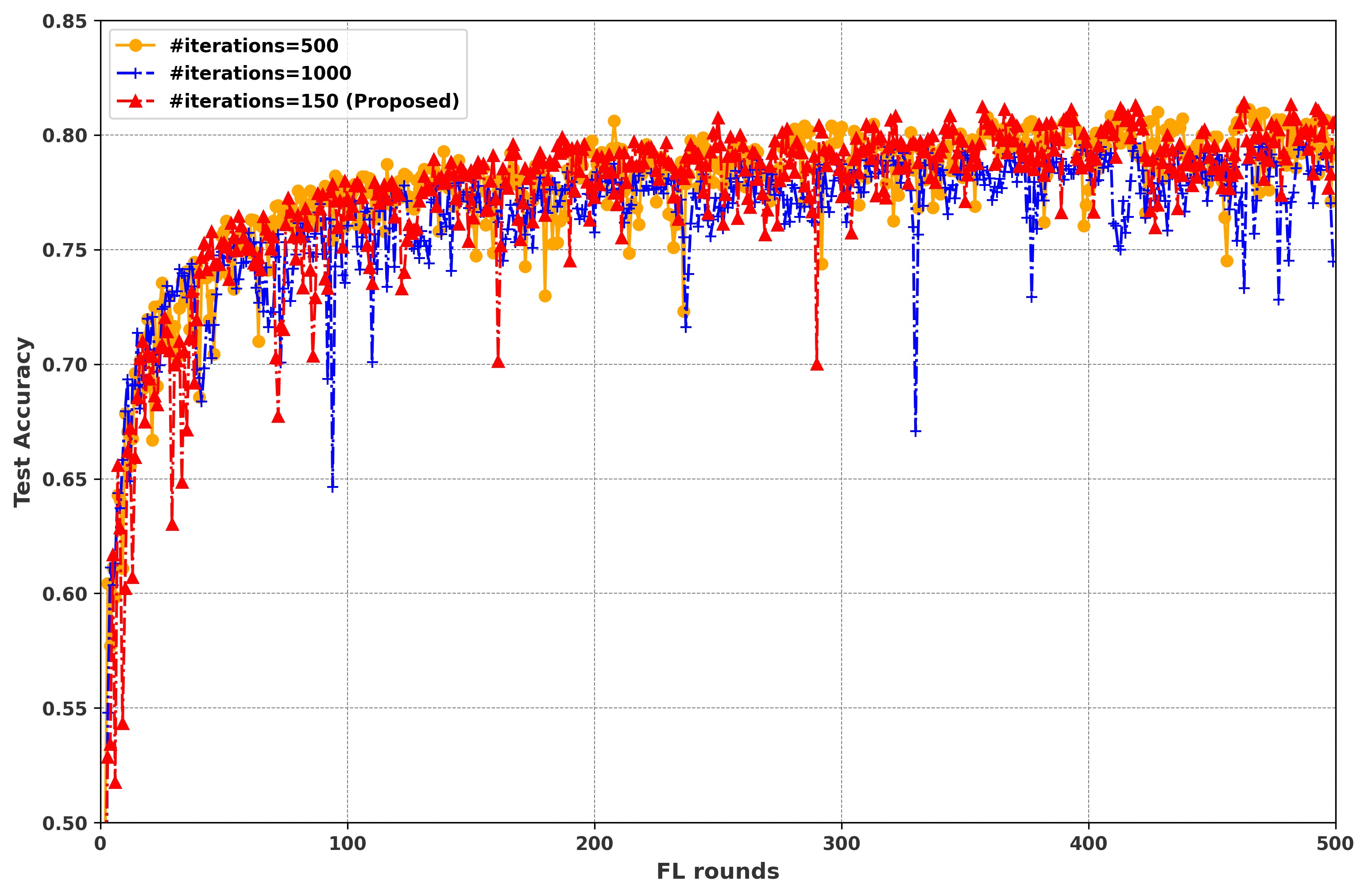}
\caption{Learning curves for different number of Iterations for WB calculation.}
\label{fig13}
\end{figure}

\begin{figure}
\centering
\includegraphics[width=0.8\linewidth]{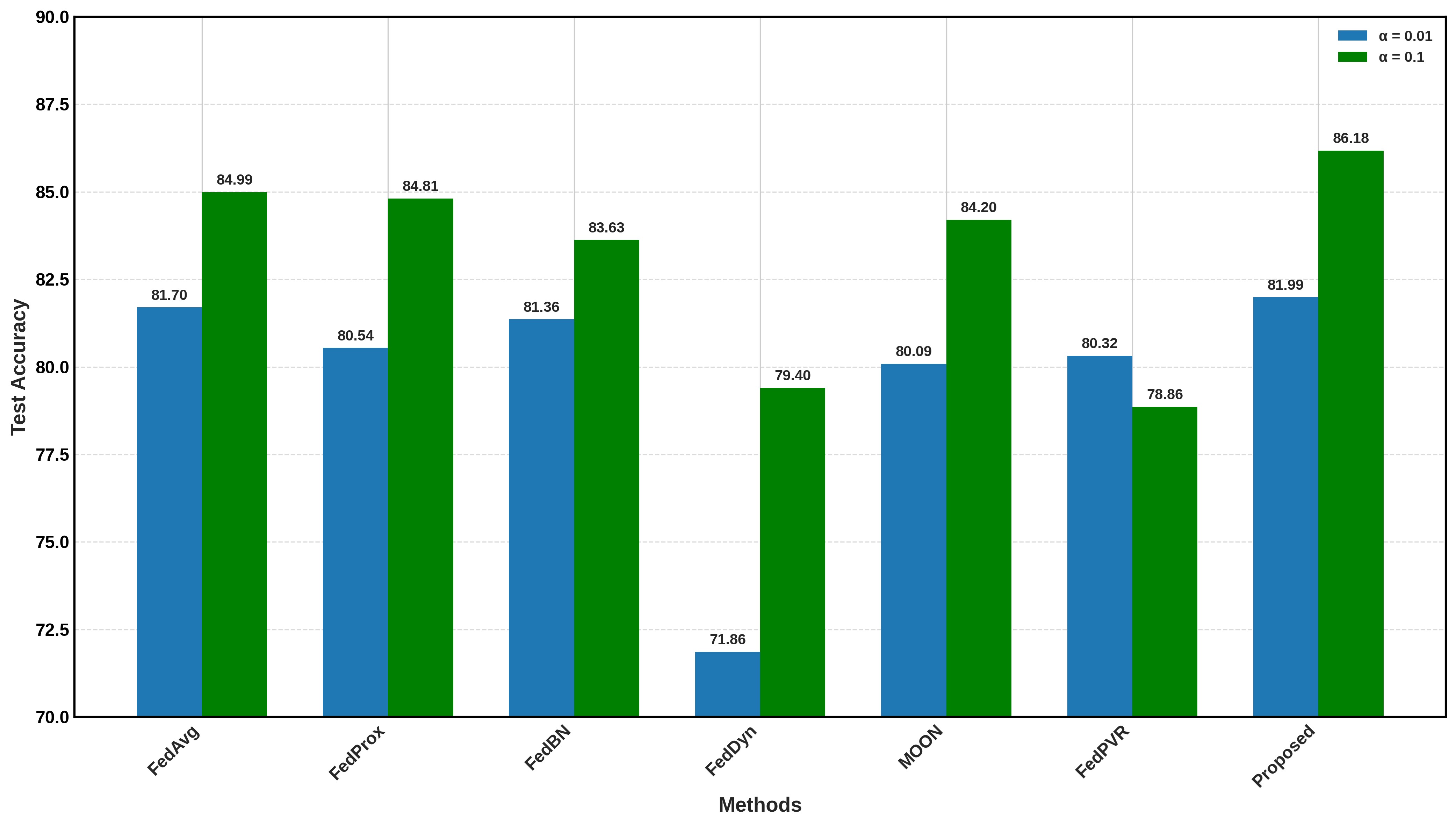}
\caption{Illustrates the accuracy of the proposed method and baselines across different levels of data heterogeneity on the FMNIST dataset.}
\label{fig131}
\end{figure}

\begin{figure}
\centering
\includegraphics[width=0.8\linewidth]{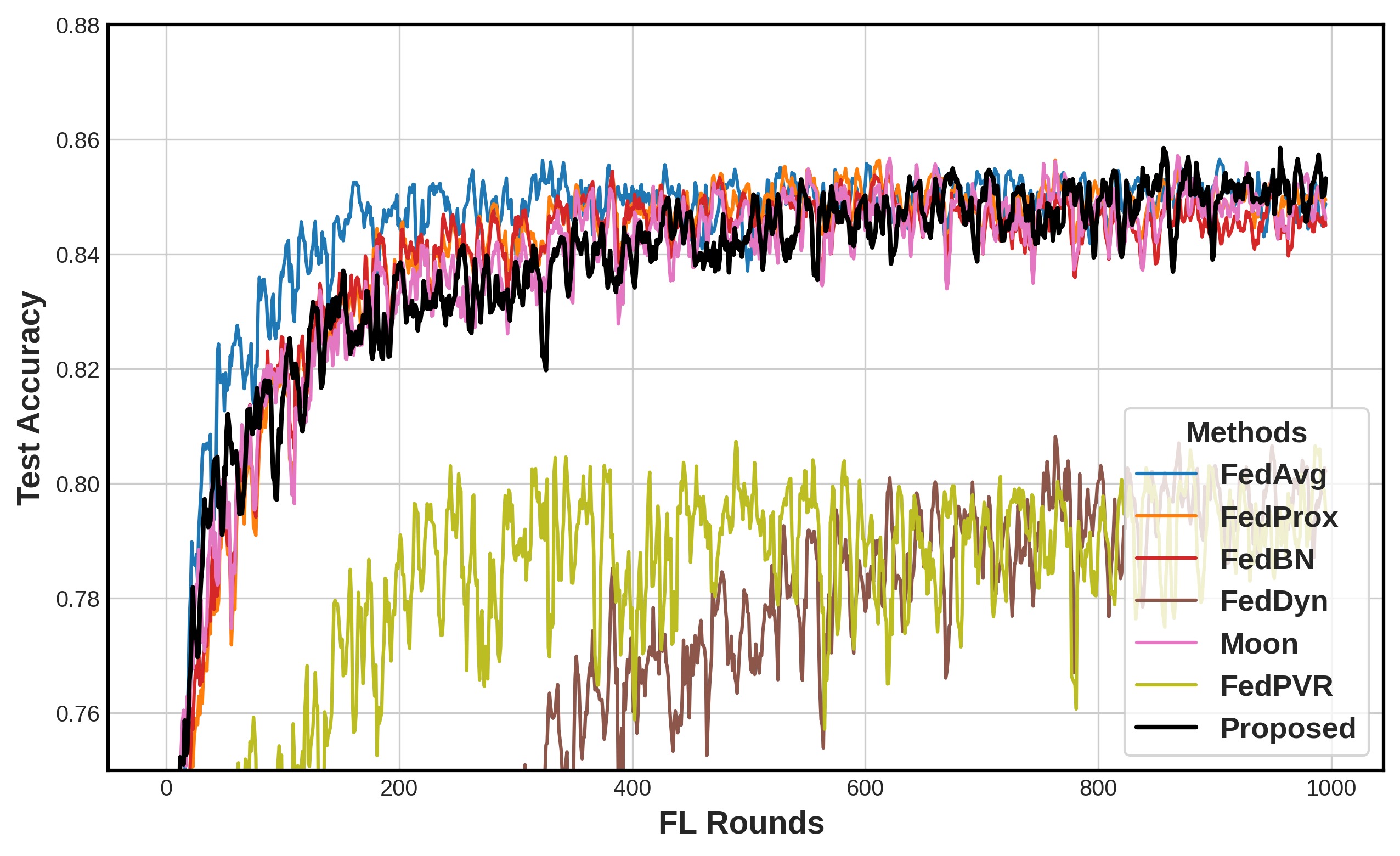}
\caption{Learning curve of the proposed method and other baselines on FMNIST dataset with data heterogeneity level $\alpha$=0.1.}
\label{1301gyg}
\end{figure}




\end{document}